\documentclass{article}

\usepackage[sglblindworkshop, final]{neurips_2025}
\workshoptitle{Foundations of Reasoning in Language Models}

\usepackage[utf8]{inputenc} % allow utf-8 input
\usepackage[T1]{fontenc}    % use 8-bit T1 fonts
\usepackage{hyperref}       % hyperlinks
\usepackage{url}            % simple URL typesetting
\usepackage{booktabs}       % professional-quality tables
\usepackage{amsfonts}       % blackboard math symbols
\usepackage{nicefrac}       % compact symbols for 1/2, etc.
\usepackage{microtype}      % microtypography
\usepackage{xcolor}         % colors
\usepackage{graphicx}
\usepackage{subcaption}
\usepackage{wrapfig}
\usepackage{amsmath}
\usepackage{array}
\usepackage{makecell}
\usepackage{booktabs}
\usepackage{multirow}
\usepackage{bbold}
\usepackage{colortbl} % For coloring table cells
\usepackage{arydshln}
\usepackage[symbol]{footmisc}
\usepackage{tcolorbox}
\usepackage{multicol} % Add this to your preamble
\usepackage{tabularx}
\usepackage{algorithm}
\usepackage{algpseudocode}
\usepackage{setspace}
\usepackage{enumitem}
\usepackage{lipsum}
\tcbuselibrary{breakable}
\usepackage{wrapfig}

\title{Grounding LLM Reasoning with Knowledge Graphs}

\author{%
  \begingroup
  \hypersetup{pdfborder={0 0 0}}% Remove border for this link only
  Alfonso Amayuelas\thanks{Work completed during an internship at JP Morgan AI Research}%
  \endgroup\\
  UC Santa Barbara\\
  \texttt{\small amayuelas@ucsb.edu} \\
  \And
  Joy Sain, Simerjot Kaur, Charese Smiley \\
  JP Morgan AI Research \\
  \texttt{\small{joy.sain, simerjot.kaur, charese.h.smiley}@jpmchase.com} \\
}

\begin{document}

\maketitle

\begin{abstract}
% Knowledge Graphs (KGs) are valuable for representing relationships between entities in a structured format. Traditionally, these knowledge bases are queried to extract specific information. However, question-answering (QA) over KGs poses a challenge due to the intrinsic complexity of natural language compared to the structured format and the vast size of these graphs. Despite these challenges, the structured nature of KGs offer a robust foundation for grounding the outputs of Large Language Models (LLMs), enhancing reliability and control for organizations. In this work, we introduce a novel integration of reasoning strategies with KGs, anchoring each step or ``thought'' of the reasoning chains in KG data. This approach uses recent advancements in LLMs, applying reasoning methods during inference to improve performance and capabilities. We evaluate both agentic and automated search methods across several reasoning strategies, including Chain-of-Thought (CoT), Tree-of-Thought (ToT), and Graph-of-Thought (GoT), using GRBench, a benchmark dataset for graph reasoning with domain-specific graphs. Our experiments demonstrate that this innovative approach achieves a significant performance improvement of at least 26.5\%  over baseline models, highlighting the benefits of grounding LLM reasoning processes in structured KG data.
Large Language Models (LLMs) excel at generating natural language answers, yet their outputs often remain unverifiable and difficult to trace. Knowledge Graphs (KGs) offer a complementary strength by representing entities and their relationships in structured form, providing a foundation for more reliable reasoning. We propose a novel framework that integrates LLM reasoning with KGs by linking each step of the reasoning process to graph-structured data. This grounding turns intermediate ``thoughts'' into interpretable traces that remain consistent with external knowledge. Our approach incorporates multiple reasoning strategies, Chain-of-Thought (CoT), Tree-of-Thought (ToT), and Graph-of-Thought (GoT), and is evaluated on GRBench, a benchmark for domain-specific graph reasoning. Our experiments show state-of-the-art (SOTA) performance, with at least 26.5\% improvement over CoT baselines. Beyond accuracy, we analyze how step depth, branching structure, and model size influence reasoning quality, offering insights into the conditions that support effective reasoning. Together, these contributions highlight how grounding LLMs in structured knowledge enables both higher accuracy and greater interpretability in complex reasoning tasks.

\end{abstract}

\section{Introduction}
\vspace{-2pt}

% Abstract + Introduction + 1 Figure should take the 1st 2 pages

LLMs have shown remarkable versatility in answering questions posed in natural language. This is mainly due to their ability to generate text, their broad internal knowledge, and their capacity to access external information \citep{toolqa, rag}.
%@Joy: Rephrase
% This is mainly due to their broad internal knowledge, their capacity to access external information \citep{toolqa, rag} and their ability to generate text.
However, a significant area for improvement is their tendency to produce information that, while plausible-sounding, is often unverifiable and lacks traceable origins and sources \citep{hallucination_survey}. This limitation highlights a deeper issue in how LLMs organize and apply reasoning, especially when reliable and accountable outputs are required.
%@Joy: Rephrase
%However, the generated responses frequently include information that may sound plausible but is often unverifiable and lacks traceable origins and sources. \citep{hallucination_survey}.  

% The LLM content generation process is based on internal parameters, making it challenging to link their outputs to external sources \citep{llm_internal_knowledge, llm_kb}. The need for more transparency in how these models generate content further complicates efforts to ensure the reliability and accountability of their responses. Consequently, this impacts their suitability for industrial applications \citep{llm_explainability_survey}.

The LLM generation process heavily relies on their internal parameters, making it difficult to link their outputs to external sources \citep{llm_internal_knowledge, llm_kb}.  This limitation challenges their reliability in industrial applications \citep{llm_explainability_survey}. In applied settings, where LLMs handle critical operations, integrating them with domain-specific knowledge is essential. Fine-tuning LLMs for new domains is labor-intensive, especially for companies with proprietary data facing privacy and legal issues. As a result, there is a need for interventions that guide reasoning processes so that outputs remain accurate and transparent without requiring exhaustive re-training.

% This lack of transparency in content generation poses challenges in ensuring the reliability and accountability of their responses. Thus, this limitation impacts their suitability for industrial applications \citep{llm_explainability_survey}.
% These shortcomings become more relevant in applied settings, where LLMs are used for critical operations within organizations. This highlights the importance of integrating LLM generation processes with domain-specific knowledge. Since LLMs are typically trained on general datasets, fine-tuning them for each new domain can be labor-intensive. This challenge is even more pronounced for companies with proprietary internal data due to privacy, legal, and resource challenges. Hence, it is crucial to develop processes that effectively link LLMs with external knowledge bases. 

Methods such as Retrieval-Augmented Generation (RAG) \citep{rag} and SQL-based querying \citep{llm_eval_sql} address this gap partially. However, they often fail to capture the dynamic relationships between concepts that are necessary for comprehensive understanding. These approaches typically assume that knowledge is well-represented in discrete units, such as documents or tables, which can lead to incomplete insights when dealing with interconnected knowledge that spans multiple sources. This limits the ability to support reasoning over complex queries.

% they typically assume that knowledge is well represented in discrete units, such as documents or tables. This approach fails to capture the complex relationships between concepts that may extend beyond one particular document. 

\begin{figure*}[t]
    \centering
    \includegraphics[width=.955\linewidth]{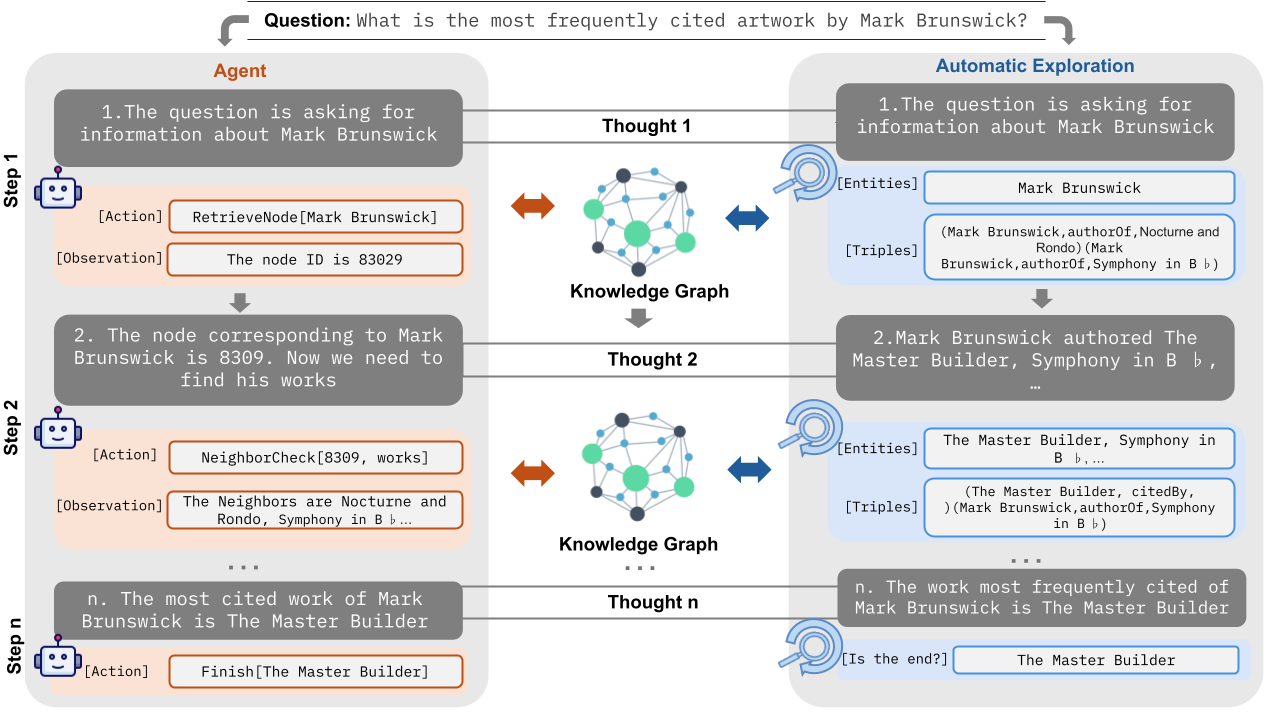}
    \caption{Methods for Question-Answering in KGs (Section \ref{Section:Method}). \textbf{Left}: \textit{Agent}. LLM decides to take one of the predefined actions to connect with the graph. \textbf{Right:} \textit{Automatic Graph Exploration}. Entities are extracted in each reasoning step, triggering a search for each identified entity.} 
    \label{fig:general}
    \vspace{-12pt}
\end{figure*}

% , explaining the general framework described in

% KGs emerge as a solution for more effectively representing complex knowledge. A KG is an organized representation of real-world entities and their relationships. This interconnected nature of concepts is reflected both in written content and in the patterns of their relationships. However, employing them to guide the reasoning processes of LLMs is a novel approach that has only recently started to be explored \citep{rog}.
% By connecting reasoning strategies with graph-structured knowledge, this work contributes both practical interventions and conceptual insights that advance the study of reasoning in language models.

KGs capture such relationships by organizing entities and their connections in a structured representation. Recent work has begun to explore how KGs can guide reasoning in LLMs \citep{rog}. Building on this direction, we propose a framework that integrates reasoning strategies with domain-specific KGs from the GRBench dataset. Each reasoning step is explicitly connected to the graph, producing answers that are both accurate and traceable. We evaluate three strategies, CoT, ToT, and GoT, which provide different ways of organizing reasoning traces (Figure \ref{fig:reasoning-strategies}). Our framework achieves SOTA performance on GRBench, improving by at least 26.5\% over CoT baselines. Beyond accuracy, we analyze how reasoning behaviors vary with step depth, branching, and model size (Section \ref{sec:Analysis}), offering insights into how reasoning evolves during inference and how it can be shaped by structured knowledge.  Our contributions can be summarized as follows:

\begin{itemize}[topsep=0pt, itemsep=0pt]
    \item We present a versatile framework that links reasoning steps with graph search, providing an intervention that can be applied across domains and knowledge settings.
    
    \item We show substantial gains over existing reasoning strategies, with state-of-the-art performance on GRBench and an improvement of at least 26.5\% compared to CoT baselines.
    
    \item We conduct a detailed study of how reasoning quality is influenced by step depth, branching structure, and model size, offering insights into the conditions under which different interventions succeed (Section \ref{sec:Analysis}).

    \item By systematically grounding each step in a knowledge graph, we improve both the accuracy and the traceability of outputs, creating reasoning traces that are interpretable and verifiable.
\end{itemize}

%This task is commonly known as Knowledge-Graph Question Answering (KGQA)
% \textcolor{red}{SK: IS THIS NEEDED? Its a repeat: Using structured knowledge, particularly Knowledge Graphs, enhances traceability and provides guarantees for organizations that use LLMs in their processes. It also allows organizations to leverage their own data better. This work introduces methods that advance reasoning capabilities on knowledge graphs and emphasizes the importance of connecting each step of the reasoning process with the KG to guide the model more effectively in the search. It has the potential to be extended to other real-world environments beyond knowledge bases, such as databases and the physical world.}

\section{Related Work}
%TODO: Either add 1 short paragraph or extend 1 of them

% LLMs require substantial data and resources for training \cite{villalobos2024will}. To extend their capabilities beyond fixed training corpora, Retrieval-Augmented Generation (RAG) methods have become prominent, enabling models to incorporate external evidence at inference time \citep{rag, demonstrate-search-predict}. Recent advances go further by combining RAG with structured knowledge, such as ontologies and KGs, to enhance factuality and reasoning capabilities \cite{li2024structrag}. This trend highlights the growing importance of structured data in improving the robustness and domain-specific adaptability of LLMs.  

LLMs require substantial data and resources for training \cite{villalobos2024will}. Retrieval-Augmented Generation (RAG) enable the models to incorporate external evidence at inference time \citep{rag, demonstrate-search-predict}. Recent work further combines RAG with structured knowledge, such as ontologies and KGs, to improve factuality and reasoning \cite{li2024structrag}, underscoring the growing importance of structured data for robust and domain-adaptable LLMs.

 % Our work introduces a framework that employs advanced reasoning strategies, grounding LLM outputs in domain-specific KGs for improved performance in specific domains.  This approach facilitates more targeted and iterative interactions with knowledge graphs, distinguishing it from traditional RAG methods.

\paragraph{Structured Knowledge} Structured knowledge, such as databases or KGs, provides organizations with reliable sources of information that can be systematically maintained and automatically updated. KGs in particular offer an adaptable knowledge model that captures complex relationships between interconnected concepts. Research has explored models that can interact with multiple types of structured knowledge, such as StructLM \citep{structLM}, and approaches that incorporate structured knowledge during pretraining to improve model performance \citep{skill_llm_structured_knowledge_infusion}. 

\paragraph{Integrating KGs with LLMs} The integration of KGs with LLMs has emerged as a promising direction to strengthen reasoning capabilities and reliability \citep{peng2024graph}. In general, four primary methods can be distinguished: (1) learning graph representations \citep{let_graph_do_thinking, graphllm}, though latent representations often underperform compared to text-based methods on Knowledge Graph Question Answering (KGQA) tasks; (2) using Graph Neural Network (GNN) retrievers to extract relevant entities and feeding them as text-based input to the model \citep{g-retriever, gnn-rag}; (3) generating code, such as SPARQL queries, to directly retrieve information from graphs \citep{kb-binder}; and (4) step-by-step interaction methods that allow iterative reasoning over graphs \citep{rog, decaf, chatkbqa}, which currently achieve the strongest results on KGQA benchmarks. 

% \paragraph{LLM Reasoning with Graphs} There is a general interest in leveraging knowledge graphs as a data structure to improve the reasoning capabilities of LLMs. Knowledge Graphs can also be used to understand the reasoning capabilities of LLMs \cite{wang2024understanding}. In most cases, they use the last form of interaction presented in the previous paragraph. In which the LLMs are allowed to interact with the graphs. Some examples can be found in \citet{think-on-graph, mindmap, rog, kg-gpt}. More recently, pointing in the same direction as our work, some works explore the connection of the traditional LLM reasoning strategies with KGs \citet{graphCoT, tree-of-traversals}.

%\paragraph{LLM Reasoning with Graphs} There is a growing interest in leveraging KGs to enhance the reasoning capabilities of LLMs. KGs not only serve as a structured data source but also provide a framework for understanding and improving the reasoning processes of LLMs \cite{wang2024understanding}. This integration enables models to generate more coherent and contextually relevant responses while allowing the tracing and verification of the reasoning steps. The most effective methods typically involve a step-by-step interaction between LLMs and graphs, as discussed in the previous paragraph. Notable examples of this approach include the works of \citet{think-on-graph, mindmap, rog, kg-gpt, li2025cot}. Recent research, including our own, has begun to explore the integration of traditional LLM reasoning strategies with KGs, as shown in studies by \citet{graphCoT, tree-of-traversals}.

\paragraph{LLM Reasoning with Graphs} Beyond retrieval, KGs have also been studied as a means to structure and analyze the reasoning processes of LLMs themselves \cite{wang2024understanding}. This integration enables more coherent and contextually relevant outputs while also supporting the tracing and verification of reasoning steps. The most effective methods typically rely on interactive, step-by-step engagement between LLMs and graphs, as discussed above. Examples of this approach include systems such as \citet{think-on-graph, mindmap, rog, kg-gpt, li2025cot}, which demonstrate improved reasoning performance through graph-based scaffolding. More recent work has further investigated the integration of traditional reasoning strategies, such as CoT and tree-structured reasoning, into KG-based interaction \citep{graphCoT, tree-of-traversals}.

Building on these advances, our framework integrates established reasoning strategies directly with domain-specific KGs. Unlike previous methods that treat KGs as retrieval tools or rely on latent representations, our approach systematically links each reasoning step to graph entities and relations. 

% This grounding enables more accurate, interpretable, and verifiable reasoning while supporting iterative exploration of the knowledge graph. 

\vspace{-4pt}
\section{Background}
\label{Section:Preliminaries}
\vspace{-4pt}

In this section, we formalize the prerequisite knowledge relevant to this paper. We use $p_\theta$ to denote a pre-trained language model with parameters $\theta$, and letters $x,y,z$ to refer to a language sequence. $x=(x_1, x_2, ..., x_n)$, where each generated token is $x_i$ is a such that $p_\theta(x) = \prod_{i=1}^n p_\theta(x_i | x_{1...i-1})$.

\paragraph{Knowledge Graphs \textnormal{(KGs)}} A KG is a heterogeneous directed graph that contains factual knowledge to model structured information. Nodes represent entities, events, or concepts, while edges represent the connection and types of relations between them.
Formally, a KG is represented as $\mathcal{G}$, defined by a set of triples $\mathcal{G} = \{ (h,r,t)  \mid h,t \in \mathcal{E}, r \in \mathcal{R} \} $, where $\mathcal{E}$, $\mathcal{R}$ denote the set of entities and relations, respectively. KGs provide a structured framework that can guide reasoning processes by explicitly representing the relationships between concepts.

\paragraph{Knowledge Graph Question-Answering \textnormal{(KGQA)}} It is a reasoning task that leverages KGs. Given a natural language question, $q$, and an associated KG, $\mathcal{G}$, the goal is to develop a method that retrieves the correct answer, $a$, based on the knowledge extracted from the KG: $a = f(q, \mathcal{G})$. Beyond retrieving facts, KGQA often requires integrating multiple reasoning steps that traverse the graph to connect related concepts.

\paragraph{Step-by-step Reasoning with LLMs} To improve the reasoning capabilities of LLMs at inference time, a common approach is to generate intermediate reasoning steps. The key idea is the introduction of intermediate steps, $Z_{p_\theta} = z_1, ... , z_n$, to add inference sources to bridge the $q$ and $a$. This decomposition allows models to tackle complex, multi-step problems incrementally, focusing computational effort on parts of reasoning chain that require deeper analysis. Stepwise reasoning over KGs offers a natural mechanism to track, guide, and interpret the reasoning process.

\section{Method}
\label{Section:Method}

This work demonstrates how progressively conditioning LLM reasoning at each step can enhance performance on domain-specific question answering over knowledge graphs. By structuring reasoning into incremental steps that interact with graph data, the model can manage complex dependencies and dynamically refine its conclusions. Our method combines reasoning strategies for LLMs: \textit{CoT, ToT, GoT} with 2 graph interaction methods: (1) Agent, an agent to navigate the graph; and (2) Automatic Graph Exploration, an automatic graph traversal mechanism based on the generated text.

\begin{figure*}[tp]
    \centering
    \begin{subfigure}[b]{0.15\textwidth}
        \centering
        \includegraphics[width=\textwidth]{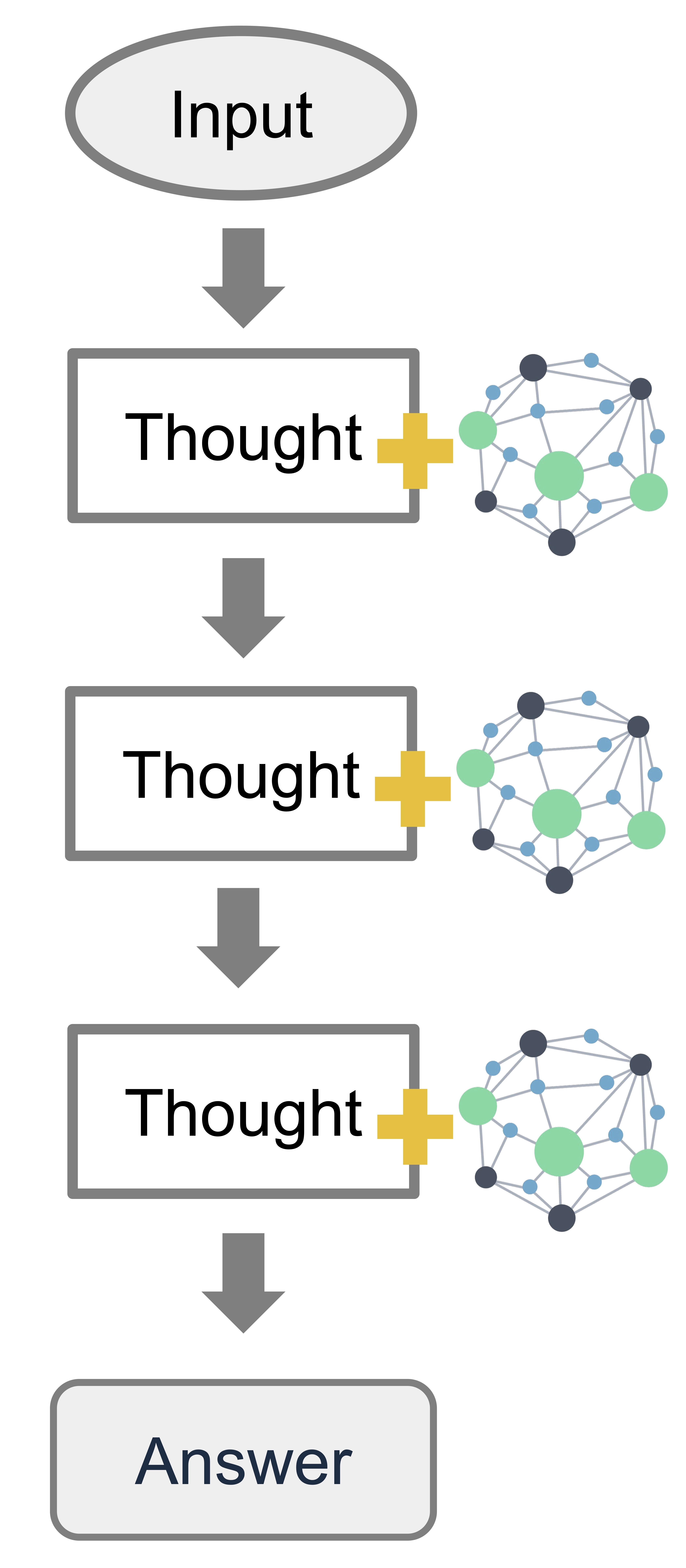}  % White box simulating a white image
        % \caption{\centering Chain of Thought (CoT)}
        \caption{\centering Chain of Thought (CoT)}
        \label{fig:graph-cot}
    \end{subfigure}
    \hfill
    \begin{subfigure}[b]{0.4\textwidth}
        \centering
        \includegraphics[width=\textwidth]{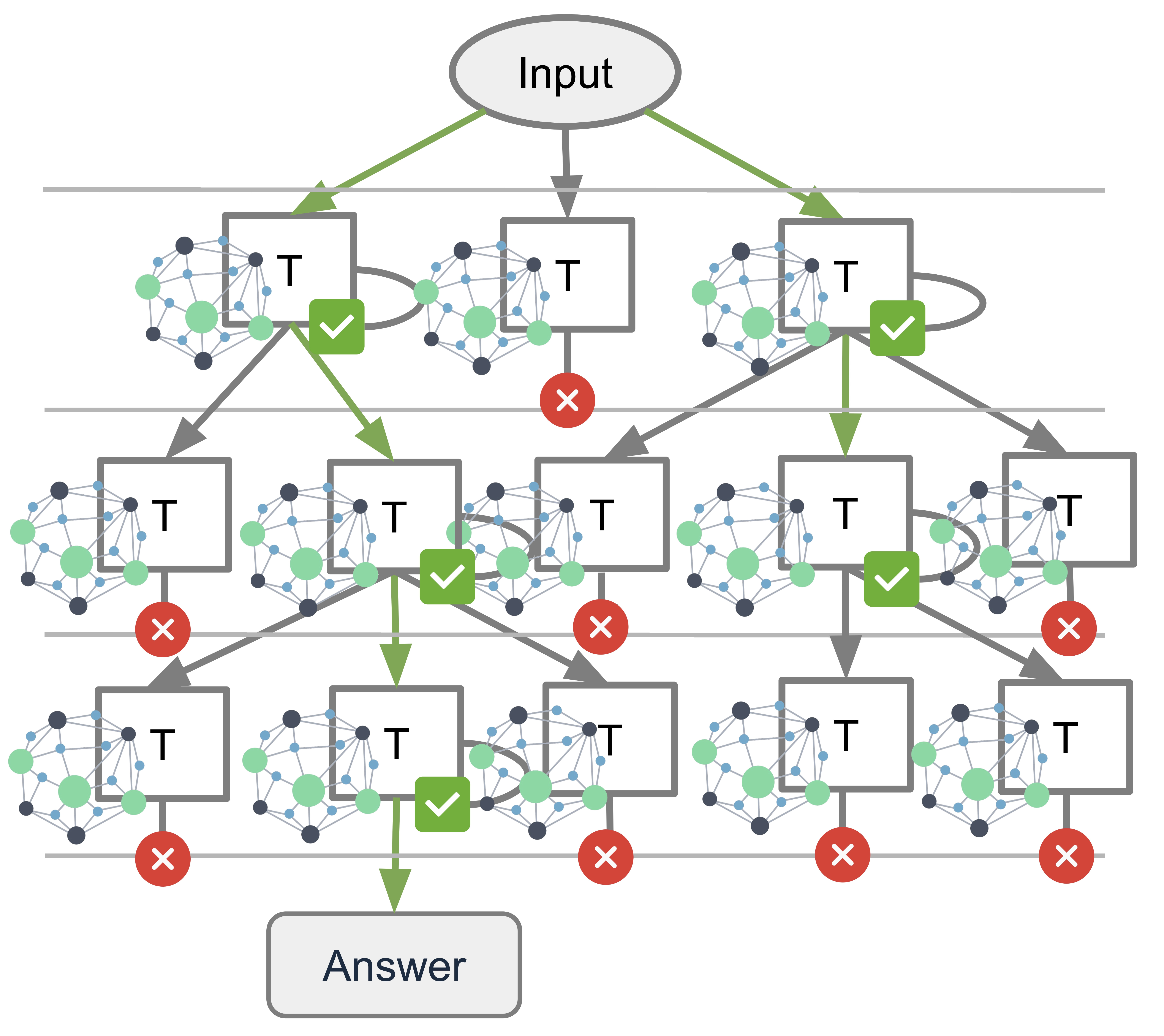} % White box simulating a white image
        % \caption{\centering Tree of Thought \\ (ToT)}
        \caption{Tree of Thought\\ (ToT)}
        \label{fig:graph-tot}
    \end{subfigure}
    \hfill
    \begin{subfigure}[b]{0.42\textwidth}
        \centering
        \includegraphics[width=\textwidth]{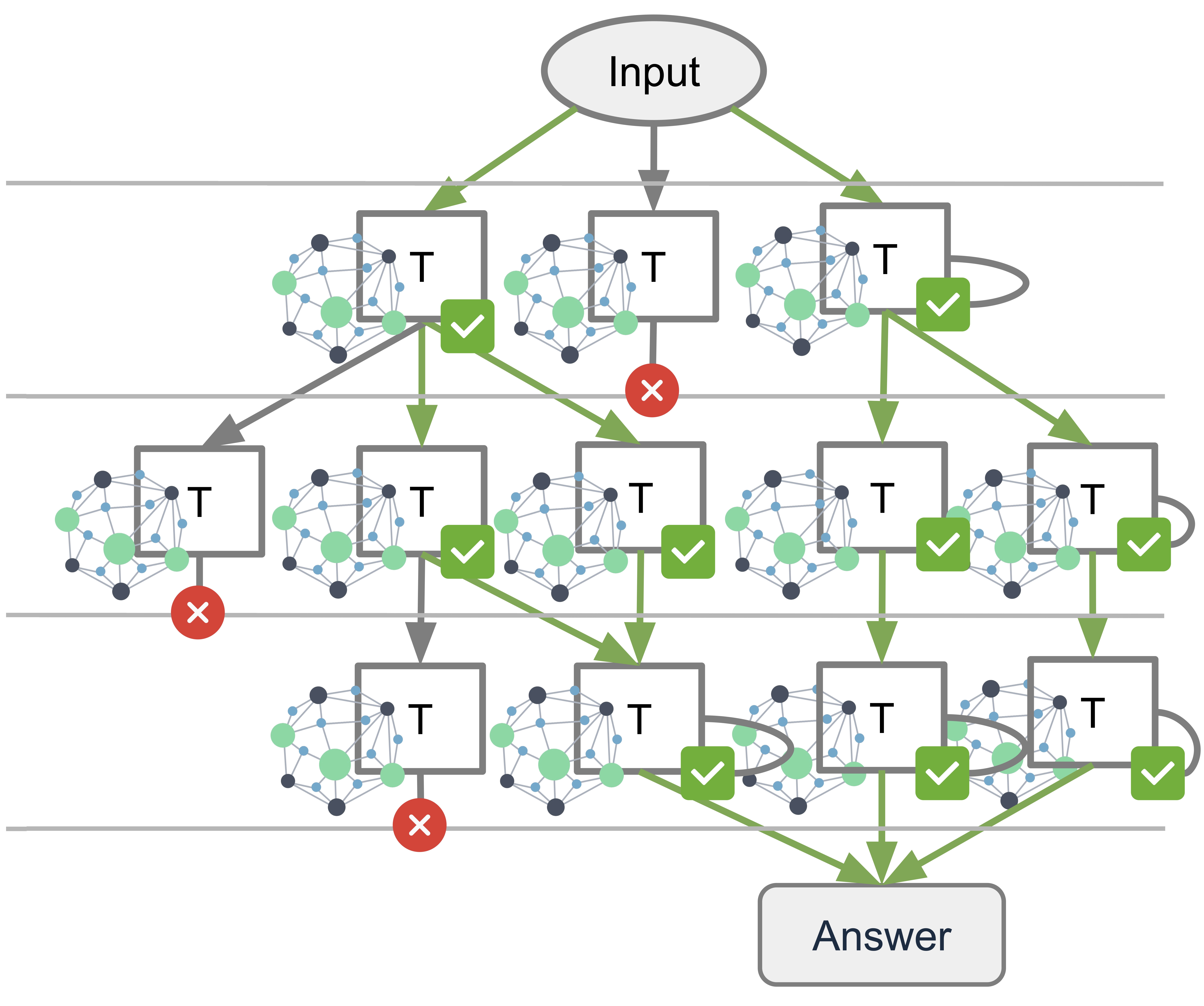} % White box simulating a white image
        % \caption{\centering Graph of Thought \\ (GoT)}
        \caption{Graph of Thought \\ (GoT)}
        \label{fig:graph-got}
    \end{subfigure}
    \caption{Reasoning Strategies: This figure illustrates different LLM reasoning strategies to navigate the potential answer space: CoT, ToT, GoT. Each strategy consists of  "thoughts" connected to the KG through search methods (Section \ref{Section:LM-KG-Interaction}) illustrating stepwise reasoning over structured knowledge.}
    \label{fig:reasoning-strategies}
\end{figure*}

\subsection{Reasoning Strategies} 

\paragraph{Chain-of-Thought (CoT)} CoT is a well-known reasoning method that involves generating a sequence of logical steps, where each step builds upon previous ones, ultimately leading to a conclusion. Formally, it generates a sequence of reasoning steps \(Z_{p_\theta} (q) = \{z_1, z_2, \ldots, z_n\}\), where each step \(z_i\) is sampled sequentially given the input query \(q\), all previous steps and graph information from all steps , $\mathcal{G}'$, as \(z_i \sim p_{\theta}^{\text{CoT}} (z_i | q, \mathcal{G}', z_{1 \ldots i-1})\). The final answer \(a\) is derived from this reasoning process given all the generated thoughts $a \sim  p_{\theta}^{\text{CoT}} (a | q, \mathcal{G}', z_{1 \ldots n})$. In practice, it is sampled as a continuous language sequence. Figure \ref{fig:graph-cot} represents this method, where each step is linked to the KG.

\paragraph{Tree-of-Thought (ToT)}

ToT generalizes CoT by modeling the reasoning process as a tree, enabling simultaneous exploration of multiple reasoning paths. Starting from an initial state $s_0 = [q]$, where $q$ is the input, ToT incrementally expands each state by generating multiple candidate thoughts:
\begin{equation}
z_{i+1}^{(j)} \sim p_\theta(z_{i+1} \mid s_i), \quad j = 1, \dots, k
\end{equation}
Each candidate thought represents a node in the tree, forming new states. These states are evaluated by a heuristic scoring function $V(p_\theta, s)$, guiding the selection and pruning of branches. Search strategies, such as breadth-first search (BFS), systematically explore this tree:
\begin{equation}
S_t = \text{argmax}_{S' \subseteq \hat{S}_t, |S'|=b} \sum_{s \in S'} V(p_\theta, s)
\end{equation}
where $\hat{S}_t$ denotes candidate states at step $t$, and $b$ limits the breadth. We implement two versions of heuristic functions $V$ to select the top $t$ states:
\begin{enumerate}
    \item  Selection: The LLM directly chooses the top $t$ states to proceed, discarding the others.
    \item Score: The states are ranked by a heuristic voting mechanism: $V(p_{\theta}, S) (s) = \mathbb{P}[s=s^*]$  where the LLM is prompted to estimate probability of the current state solving the given input question.
\end{enumerate}

This structured search and pruning strategy allows the model to evaluate multiple candidate reasoning paths, enabling more deliberate and interpretable reasoning.

\paragraph{Graph-of-Thought (GoT)} 

GoT extends ToT by organizing reasoning into a directed graph structure $G = (V, E)$, where each node represents a thought and edges reflect dependencies. Starting from initial thought, new thoughts are generated similarly to ToT and added to the graph. Each new thought is connected to its parent, and additional reasoning chains can be formed through merging operations:
\begin{equation}
z_{i+1} = A(z_i^{(a)}, z_i^{(b)})
\end{equation}
where $A$ denotes a merge operation that integrates two thought chains into a single coherent reasoning step. The merged thought is added as a new node with edges from both parents. In our implementation, thoughts are evaluated using either Selection- or Score-based strategy as in ToT. Merged thoughts inherit information from both parents and can enhance robustness. At each depth, a fixed number of thoughts are retained using breadth-first traversal and evaluated for progression. 
The graph-based organization captures dependencies and merges information from multiple reasoning chains, supporting dynamic refinement and structured exploration of the reasoning space.

\subsection{LLM + KG Interaction Methods}
\label{Section:LM-KG-Interaction}

We implement methods to connect reasoning strategies with KGs. The LLM interacts with the KG at every step. This retrieves new information and conditions the model for subsequent steps. We present 2 methods to achieve this interaction, both illustrated in Appendix \ref{App:pipelines}.

\subsubsection{Agent}

This approach creates an agent that interacts with the graph, following the methodology initially described in ReACT \citep{react}. After generating a thought, the LLM selects from a set of actions based on the given thought. Each step in the reasoning chain consists of an interleaved sequence: \textit{thought} $\rightarrow$ \textit{action} $\rightarrow$ \textit{retrieved data}. This method implements four actions as described in GraphCoT \citep{graphCoT}: (a) \texttt{RetrieveNode}(Text): Identifies the related node in the graph using semantic search, (b) \texttt{NodeFeature}(NodeID, FeatureName): Retrieves textual information for a specific node from the graph, (c) \texttt{NeighborCheck}(NodeID, EdgeType): Retrieves neighbors' information for a specific node, (d) \texttt{NodeDegree}(NodeID, EdgeType): Returns the degree (\#neighbors) for a given node and edge type. These actions collectively enable the agent to navigate and extract meaningful information from the graph, enhancing the reasoning capabilities of the LLM by grounding its thoughts in structured, retrievable data. 

\subsubsection{Automatic Graph Exploration}
\label{Section:Automatic_Graph_Exploration}

This method incrementally searches the graph by interleaving language generation with structured retrieval. At each step, the LLM generates a new "thought" based on previous thoughts and retrieved triples. Entities mentioned in the generated text are automatically extracted using LLM prompts and serve as anchors for further graph exploration.

% This method incrementally searches the graph by interleaving language generation with structured graph retrieval. At each step, the LLM generates a new "thought" based on previous thoughts and retrieved triples. Entities mentioned in the generated text are automatically extracted using LLM-based prompts and serve as anchors for further graph exploration.

\begin{wrapfigure}{b}{0.5\textwidth}
% \vspace{-14pt}
    \centering
    \includegraphics[width=\linewidth]{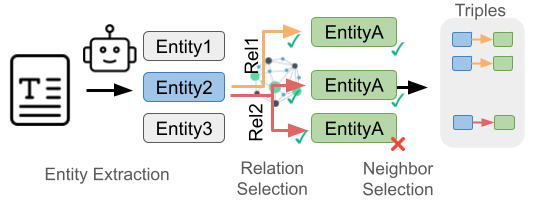}
    \caption{Automatic Graph Exploration. It extracts entities from text (query/thought), then select relevant relations and neighbors with the LLM. The resulting entity-relation-entity combinations form triples to expand the reasoning chain.}
    \label{fig:graph_search}
    % \vspace{-15pt}
\end{wrapfigure}

Graph exploration proceeds through a multi-step Search + Prune pipeline, inspired by the process described in \citet{think-on-graph}. For each unvisited entity, the system first retrieves and prunes relation types using LLM guidance. Then, for each selected relation, neighboring entities are discovered and filtered using a second round of pruning. The model selects only the most relevant neighbors based on their contextual fit with the question and previous reasoning steps. This hierarchical pruning -- first on relations, then on entities -- ensures the method remains computationally tractable while preserving interpretability. The overall traversal follows a breadth-first search (BFS) pattern, with pruning decisions at each level directed by LLM. This process is shown in Figure \ref{fig:graph_search}. This iterative reasoning and retrieval process allows the model to condition future steps on progressively relevant subgraphs, shaping the reasoning trajectory. Unlike agentic methods that rely on predefined actions, the automatic approach operates in the graph space guided by the natural language, providing more freedom in the generation. The mechanism is designed to maximize information gain at each step while avoiding graph overgrowth. More details are provided in Algorithm \ref{alg:graph_exploration_algorithm}.

\section{Experiments}
\label{Section:Experiments}
\vspace{4pt}

\paragraph{Benchmark}
We use the GRBench dataset to evaluate our methods. This dataset is specifically designed to evaluate how effectively LLMs can perform stepwise reasoning over domain-specific graphs. It includes several graphs spanning various general domains. For our evaluation, we selected 7 graphs across multiple domains, excluding those with excessively high RAM requirements that exceed our available resources. Comprehensive graph statistics are provided in Appendix \ref{app:grbench_statistics}.

\paragraph{Baselines}
The proposed methods, \textit{Agent} and \textit{Automatic Graph Exploration}, applied to CoT, ToT, and GoT, are compared against the following baseline methods: (1) \textbf{Zero-Shot}: Directly querying the model to answer the question without additional context. (2) \textbf{Text RAG} \cite{rag_sruvey}: Text-retrieval method that uses text representation of nodes as input for query, with retrieved data serving as context for the model. (3) \textbf{Graph RAG}: Includes node neighbors (1-hop) for additional context beyond Text RAG. (4) \textbf{Graph CoT (Agent)}: Implements Graph CoT as an agent for CoT reasoning, utilizing the actions described in Section \ref{Section:LM-KG-Interaction}. These baselines allow us to measure impact of stepwise, knowledge-grounded reasoning versus simple retrieval-augmented or zero-shot approaches.

\paragraph{Experimental methods}
We implement the methods described in Section \ref{Section:Method}, extending (1) \textbf{Agent} and (2) \textbf{Automatic Graph Exploration} with various reasoning strategies during inference: (1) \textbf{CoT}, (2) \textbf{ToT}, and (3) \textbf{GoT}.  For ToT and GoT, we evaluate the impact of stepwise decision-making using State Evaluation methods: (1) \textit{Selection} and (2) \textit{Score}. In the results presented in Table \ref{Table:Results}, we set $n=10$ steps for all methods. ToT and GoT use a branching factor and Selection of $k = t = 3$. Our experiments focus on the effect of structured reasoning interventions on performance and stepwise refinement of answers. We use only open-access Llama 3.1 (Instruct) \cite{llama3models} as the backend models, which enhances reproducibility and allows for unlimited free calls. Specifically, we employ the 8B, 70B, and 405B versions, using the FP8 variant for the 405B model.

\paragraph{Evaluation} We use rule-based and model-based metrics to evaluate the models, following GRBench paper \cite{graphCoT}. For the rule-based metric, we use Rouge-L (R-L) \citep{rouge_metric}, which measures the longest sequence of words appearing in same order in both generated text and ground truth answer. For model-based metric, we prompt GPT-4o to assess if the model's output matches ground truth answer. GPT4Score is percentage of answers that GPT-4o identifies as correct. These evaluation methods capture not only final answer accuracy but also the fidelity of reasoning steps, reflecting the effectiveness of our interventions in guiding LLM reasoning over structured knowledge.

\paragraph{Implementation Details}
The experiments are run on NVIDIA TITAN RTX or NVIDIA A100 using Python 3.8. The models are deployed with vLLM \cite{vllm}, a memory-efficient library for LLM inference and serving. For the baseline, Mpnet-v2 is used as the retriever, and FAISS \cite{faiss} is employed for indexing.

\section{Results}
\label{Section:Labels}
\vspace{4pt}

The main results from both the baselines and experimental methods, evaluated using R-L, are presented in Table \ref{Table:Results}. For brevity, additional results using GPT4Score can be found in Appendix \ref{App:results-gpt4score}. Together, these findings allow us to compare different forms of reasoning interventions, agentic action selection, automatic graph exploration, and structured multi-path search, on their ability to guide LLMs toward accurate answers. We highlight three key insights from the findings: (1) The agentic method generally outperformed automatic graph exploration, indicating that targeted interventions on knowledge graph traversal enhance answer accuracy. (2) The ToT strategy demonstrated superior performance by effectively exploring multiple reasoning paths, showcasing the benefits of inference-time interventions that diversify reasoning trajectories. (3) Although GoT strategy showed potential, it did not significantly outperform ToT, suggesting that merging divergent reasoning paths remains a challenging intervention design problem. These results show the importance of reasoning strategies in enabling LLMs to navigate multiple paths in the graph, while also illustrating the limits of current intervention techniques.

\begin{table*}[t]
\centering
\resizebox{\textwidth}{!}{
\begin{tabular}{llllccccccc}
\toprule
& \textbf{Method} & & \multicolumn{1}{c}{\textbf{Model}} & \textbf{Healthcare} & \textbf{Goodreads} & \textbf{Biology} & \textbf{Chemistry} & \textbf{Materials Science} & \textbf{Medicine} & \textbf{Physics} \\
                  \hline
\multirow{9}{*}{\rotatebox{90}{\shortstack{Baselines}}} & \multicolumn{2}{c}{}                  & Llama 3.1 8B-Ins                   & 7.32                                & 6.18             & 10.68            & 11.69              & 8.95                      & 8.77            & 11.52            \\
& \multicolumn{2}{l}{Base}                  & Llama 3.1 70B-Ins                   & 9.74                                 & 9.79               & 11.49            & 12.58              & 10.40                       & 12.21             & 12.61            \\
& \multicolumn{2}{l}{}                  & Llama 3.1 405B-Ins                   & 8.66                               & 12.49              & 10.52           & 13.51              & 11.73                     & 11.82            & 11.63            \\

\cdashline{2-11}
& \multicolumn{2}{l}{}                  & Llama 3.1 8B-Ins                   & 8.24                               & 14.69              & 12.43            & 11.42              & 9.46                      & 10.75             & 11.29            \\
                  & \multicolumn{2}{l}{Text-RAG}                  & Llama 3.1 70B-Ins                   & 10.32                                & 18.81              & 11.87            & 16.35             & 12.25                      & 12.77             & 12.54            \\
                  & \multicolumn{2}{l}{}                  & Llama 3.1 405B-Ins           & 11.61                                & 16.23              & 16.11           & 13.82              & 14.23                      & 15.16           & 16.32            \\ \cdashline{2-11}
                  & \multicolumn{2}{l}{}                  & Llama 3.1 8B-Ins                   & 12.94                               & 22.30             & 30.72               & 34.46              & 30.20                       & 25.81             & 33.49            \\
                  & \multicolumn{2}{l}{Graph-RAG}                  & Llama 3.1 70B-Ins                   & 17.95                                & 25.36              & 38.88               & 40.90              & 41.09                      & 31.43             & 39.75            \\
                  & \multicolumn{2}{l}{}                  & Llama 3.1 405B-Ins         & 16.12                                & 23.13             & 37.57               & 42.58              & 37.74                       & 33.34             & 40.98 \\
                  \hline \hline
 \multirow{6}{*}{\rotatebox{90}{\shortstack{Graph\\CoT}}} & \multicolumn{2}{c}{}                  & Llama 3.1 8B-Ins                   & 16.83                                & 30.91               & 20.15            & 18.43               & 26.29                      & 14.95            & 21.41            \\
 & \multicolumn{2}{l}{Agent}                  & Llama 3.1 70B-Ins                   & 33.48                               & 40.98              & 50.00            & 51.53              & 49.6                      & 48.27            & 44.35           \\ 
 & \multicolumn{2}{l}{}                  & Llama 3.1 405B-Ins                  & 28.41                              & 36.56              & 41.35            & 48.36             & 47.81                     & 42.54            & 35.24           \\ \cdashline{2-11}
                  & \multicolumn{2}{l}{ \multirow{3}{*}{\shortstack{Graph\\Explore}}}                  & Llama 3.1 8B-Ins                   & 25.58                                & 32.34              & 36.65            & 35.33              & 31.06                      & 31.05                & 35.96            \\
                & \multicolumn{2}{l}{}                  & Llama 3.1 70B-Ins                   & 29.41                                & 29.60            & 44.63          & 49.49             & 39.23                     & 38.87             & 45.52            \\
                & \multicolumn{2}{l}{}                  & Llama 3.1 405B-Ins                   & 28.45                               & 43.06            & 36.93       & 38.71              & 47.49                      & 55.66                & 32.73            \\
                  \hline \hline
                  
\multirow{12}{*}{\rotatebox{90}{\shortstack{Graph\\ToT}}} & \multirow{6}{*}{Agent} &    \multirow{3}{*}{Score}          & Llama 3.1 8B-Ins                  & 28.91                                & 52.25              & 43.81            & 44.18               & 43.49                      & 36.07             & 39.56            \\
& &              & Llama 3.1 70B-Ins                  & 38.51                                & 51.58              & 64.44            & 61.13               & 55.19                      & 63.00             & 55.33            \\
& &              & Llama 3.1 405B-Ins                  & \textbf{47.51 }                               & 50.73              & \textbf{70.34}           & 64.9               & 49.02                      & \textbf{65.40}             & 44.63            \\ \cdashline{3-11}

    &                   &        \multirow{3}{*}{Select}           & Llama 3.1 8B-Ins                   & 28.67                                 & 50.59               & 42.33            & 37.07              & 40.81                      & 33.17             & 36.50            \\
    &                   &                  & Llama 3.1 70B-Ins                   & 40.26                                 & \textbf{52.59}               & 64.53            & 66.84              & 61.42                      & 61.21             & 55.89            \\

                  &                   &                   & Llama 3.1 405B-Ins                   & 46.90                                 & 51.68               & 70.27            & \textbf{67.95}              & \textbf{63.74 }                     & 64.23             & \textbf{59.56}        \\  \cdashline{2-11}
                  & \multirow{6}{*}{\shortstack{Graph\\Explore}} &      \multirow{3}{*}{Score}             & Llama 3.1 8B-Ins                   & 24.49                                & 36.80              & 35.81            & 36.41              & 34.28                      & 34.49             & 37.69            \\ 
                  &                   &                   & Llama 3.1 70B-Ins                   & 32.79                                & 38.19              & 53.83            & 58.25              & 48.55                      & 52.18                & 48.07           \\
                  &                   &                   & Llama 3.1 405B-Ins                   & 33.90                                & 42.68              & 46.87            & 57.43              & 50.46                      & 55.56                & 48.73           \\ \cdashline{3-11}

                  &                   &     \multirow{3}{*}{Select}              & Llama 3.1 8B-Ins                   & 25.04                                & 37.8              & 36.34            & 38.5              & 32.44                      & 33.31                & 34.85           \\
                  &                   &                   & Llama 3.1 70B-Ins                   & 33.40                                & 39.13              & 54.78            & 58.53              & 47.19   & 51.13                & 47.51           \\
                  &                   &                   & Llama 3.1 405B-Ins                   & 33.82                                & 43.63              & 44.47            & 59.06              & 48.52                      & 55.62                & 46.07           \\
\bottomrule
\end{tabular}
}
\caption{Rouge-L (R-L) performance results on GRBench, comparing standard LLMs, Text-RAG, Graph-RAG, Graph-CoT, and Graph-ToT. Experiments are described in Section \ref{Section:Experiments}, using LLama 3.1 - Instruct backbone models with sizes 8B, 70B, and 405B-FP8.}
% \vspace{-4pt}
\label{Table:Results}
\end{table*}

\paragraph{Agent vs Graph Search} In our experimental results, the agentic method outperformed graph exploration approach across most datasets and reasoning strategies. The agent-based method, which involves LLM selecting specific actions to interact with KG, consistently improves performance as the number of reasoning steps increases, as shown in Section \ref{sec:Analysis}. This highlights that explicit, model-driven interventions are more effective than passive expansion strategies, as they promote iterative refinement and selective focus. While graph exploration can quickly provide broad coverage, the agentic method’s targeted, stepwise interactions yield more accurate and comprehensive answers over longer sequences of reasoning.

\paragraph{Tree of Thought (ToT)}
The ToT reasoning strategy showed superior performance across its various interaction methods and state evaluators, as summarized in Table \ref{Table:Results}. ToT achieved performance improvements of 54.74\% in agent performance and 11.74\% in exploration mode compared to the CoT baseline. However, this improvement comes with the trade-off of increased inference time, highlighting the cost of inference-time reasoning interventions. The success of ToT illustrates how branching interventions that explore multiple candidate paths can substantially enhance reasoning accuracy, especially when coupled with evaluators that prune unpromising trajectories. We also compared the two State Evaluation methods (Selection and Score), finding complementary benefits depending on dataset and scale.

\begin{wraptable}{r}{0.4\textwidth}
\vspace{-10pt}
\centering
\resizebox{.45\textwidth}{!}{
\begin{tabular}{cclcc}
\toprule
\multicolumn{2}{c}{\textbf{Method}}                        & \multicolumn{1}{c}{\textbf{Model}}             & \textbf{Healthcare} & \textbf{Biology} \\
\hline
\multirow{6}{*}{\rotatebox{90}{Agent}}  & \multirow{3}{*}{\rotatebox{90}{Score}}  & Llama 3.1 8B-Ins   & 29.11           &    33.25                                                     \\
                        &                         & Llama 3.1 70B-Ins  &   30.88         &    56.64                                                     \\
                        &                         & Llama 3.1 405B-Ins &    43.53        &   48.1                                                      \\
                        \cdashline{2-5}
                        & \multirow{3}{*}{\rotatebox{90}{Select}} & Llama 3.1 8B-Ins   &  29.05          & 40.37                                                         \\
                        &                         & Llama 3.1 70B-Ins  &  40.74          &     65.59                                                    \\
                        &                         & Llama 3.1 405B-Ins &   \textbf{47.63}        &   \textbf{71.49 }                                                     \\
                        \hline
\multirow{6}{*}{\rotatebox{90}{Graph Explore}} & \multirow{3}{*}{\rotatebox{90}{Score}}  & Llama 3.1 8B-Ins   &   24.96         &      21.72                                                   \\
                        &                         & Llama 3.1 70B-Ins  &   31.24         &     50.70                                                    \\
                        &                         & Llama 3.1 405B-Ins & 35.00        &   39.10                                                      \\
                        \cdashline{2-5}
                        & \multirow{3}{*}{\rotatebox{90}{Select}} & Llama 3.1 8B-Ins   &  25.06          &   21.84                                                      \\
                        &                         & Llama 3.1 70B-Ins  &   36.95         &      52.32                                                   \\
                        &                         & Llama 3.1 405B-Ins &   33.74         &   54.64        
                        \\
                        \bottomrule
\end{tabular}
}
\caption{Graph-GoT results on GRBench using Rouge-L}
\label{Table:Results_GoT}
\end{wraptable}

 % with Llama 3.1 Instruct - 8B, 70B and 405B

\paragraph{Graph of Thought (GoT)} 
The results for GoT strategy are summarized in Table \ref{Table:Results_GoT}.  Due to the additional computational cost, we report results for two datasets only. GoT did not outperform ToT. Our initial hypothesis was that LLMs could integrate divergent results from multiple branches, but in practice the models struggled to merge these effectively. Specifically, in the graph exploration setting, models often failed to combine different triples found in separate branches. This reveals a current limitation of reasoning interventions based on aggregation: while branching helps discover diverse facts, robust mechanisms for synthesis and reconciliation are still underdeveloped. This finding opens a direction for future research into more advanced intervention strategies for merging partial reasoning outcomes.

% \usepackage{multirow}

\section{Analysis \& Ablation studies}
\label{sec:Analysis}

In this section, we want to better understand the nuances of our methods for LLM and KG grounding. We conduct an analysis on the Academic datasets from the benchmark, as they all contain the same number of samples and feature questions generated from similar templates to ensure a controlled comparison.

\begin{figure}[t]
    \centering
    \begin{minipage}{0.48\textwidth}
        \centering
        \includegraphics[width=0.8\linewidth]{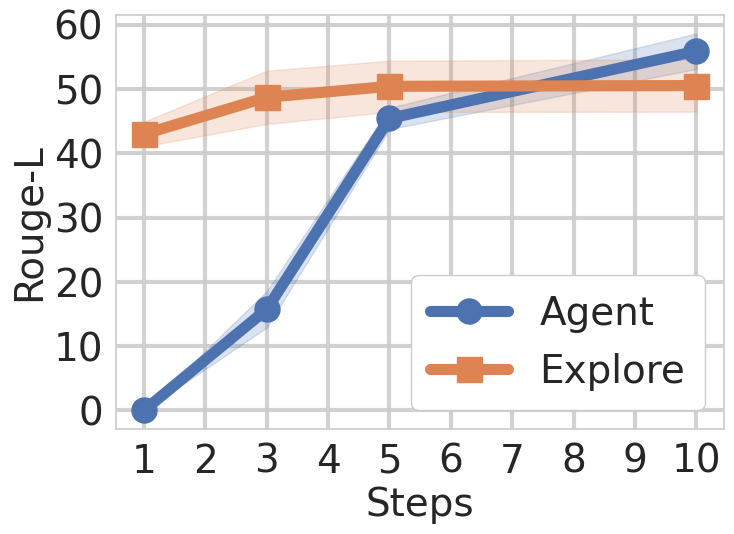}
        \caption{Effect of the number of steps in the LLM-KG Interaction Methods. The Agent requires more steps to obtain the performance of the Graph Exploration, while the Graph Exploration only needs the anchor entities to perform the search within the graph.}
        \label{fig:analysis-steps}
    \end{minipage}
    \hfill
    \begin{minipage}{0.48\textwidth}
        \centering
        \includegraphics[width=0.8\linewidth]{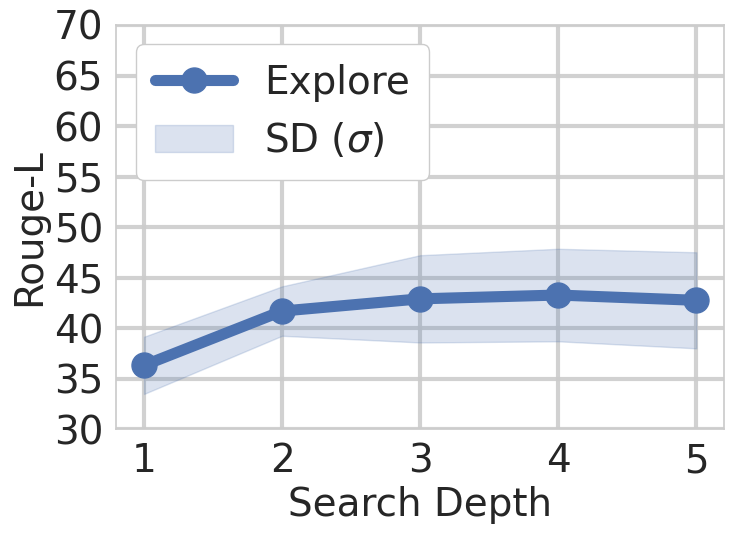} % replace with your image file
        \caption{Effect of the Search depth in Graph Exploration interaction method for a fixed steps number. The method can achieve relatively good performance with the anchor entities extracted from the question.}
    \label{fig:search-depth}
    \end{minipage}
\end{figure}

\paragraph{How does the number of steps affect the results?}
We observe in Figure \ref{fig:analysis-steps} the effect of varying the number of steps in the KG interaction methods (Agent, Explore) across all academic datasets. The plots indicate that graph exploration performs better with fewer steps, as it automatically traverses the graph for the identified anchor entities. Conversely, the agentic methods improve as the number of steps increases, eventually achieving better performance. This validates our framework’s design choice to support both exploration and agentic strategies, each excels in complementary regimes.

\begin{figure}[b]
    \centering
    \begin{minipage}{0.48\textwidth}
        \centering
        \includegraphics[width=0.8\linewidth]{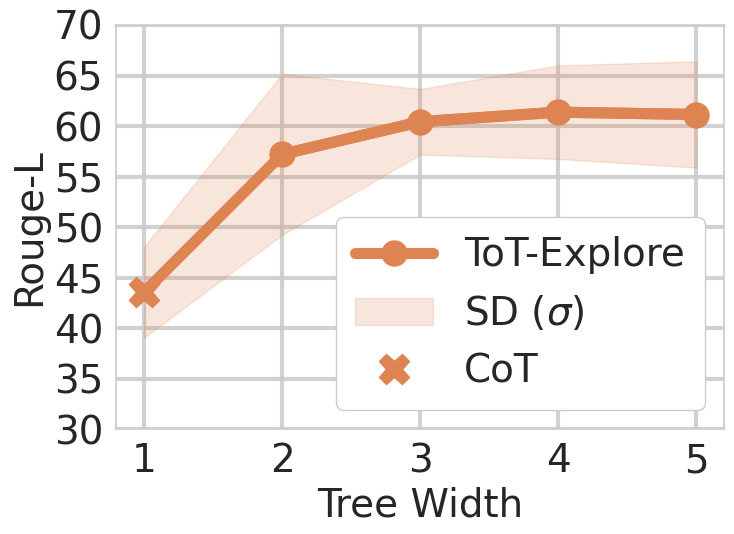}
    \caption{Impact of tree width on Agentic ToT performance. It shows a general trend of performance improvement with increasing tree width.}
    \label{fig:tree-width}
    \end{minipage}
    \hfill
    \begin{minipage}{0.48\textwidth}
        \centering
        \includegraphics[width=0.85\linewidth]{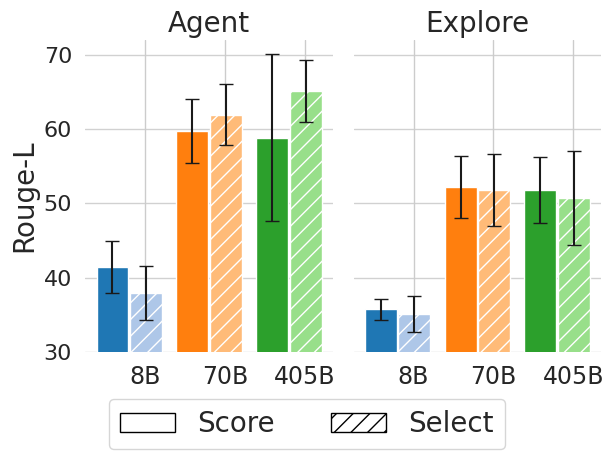}
        \caption{Influence of the State Evaluators in ToT. The Select method obtains better results over Score method.}
    \label{fig:state_evaluators}
    \end{minipage}
\end{figure}

\paragraph{What is the effect of Search Depth in Automatic Graph Exploration?}
We observe the effect of search depth in Figure \ref{fig:search-depth}, which presents performance results across various depths, with fixed step size of one. The results demonstrate that the performance of depth-first search plateaus at depth of 3, highlighting the relevance of search exploration with respect to the given query. Beyond this point, deeper traversal yields no significant gains, likely due to diminishing relevance of distant nodes. This shows why shallow, targeted exploration is sufficient in our framework, keeping search efficient without sacrificing accuracy.

\paragraph{What is the effect of tree width in the reasoning strategy (ToT)?}
Based on experimental results across all academic datasets, we observe performance variations among different methods. To gain further insight, we observe in Figure \ref{fig:tree-width} the effect of tree width on results. We notice a slight upward trend in performance as the tree width increases, although the difference is more pronounced between CoT and ToT itself, going from one branch to two. The added computational time and resources likely contribute to this performance enhancement.

\paragraph{What is the influence of the state evaluator?}
We observe in Figure \ref{fig:state_evaluators} the impact of state evaluators, specifically Score and Select, within the ToT framework. The analysis indicates that, while there is no significant difference between the two methods, the Select evaluator generally yields slightly better results. This trend is especially evident in the context of the Agent's performance, though the advantage is less pronounced in automatic graph exploration.

\paragraph{How are errors different for each strategy?}

% To understand the failure patterns of each reasoning approach, we categorize errors into three types: (1) Reached limit — the reasoning reached the limit nuber of steps; (2) Answer found but not returned — the correct answer was identified in intermediate reasoning but not produced in the final output; (3) Wrong reasoning step — the model followed an incorrect logical path. We prompted GPT-4o with each generated answer and its associated reasoning trace to classify a larger set of outputs. As shown in Figure~\ref{fig:errors_pie}, ToT and GoT strategies result in a higher proportion of "answer found but not returned" cases compared to CoT, indicating improved retrieval but occasional failures in answer synthesis. This gain comes with a slight increase in logical errors, possibly due to the increased complexity from sampling and managing multiple reasoning paths.

\begin{wrapfigure}{r}{0.5\textwidth}
\vspace{-12pt}
    \centering
    \begin{subfigure}[b]{0.28\linewidth}
        \includegraphics[width=\textwidth]{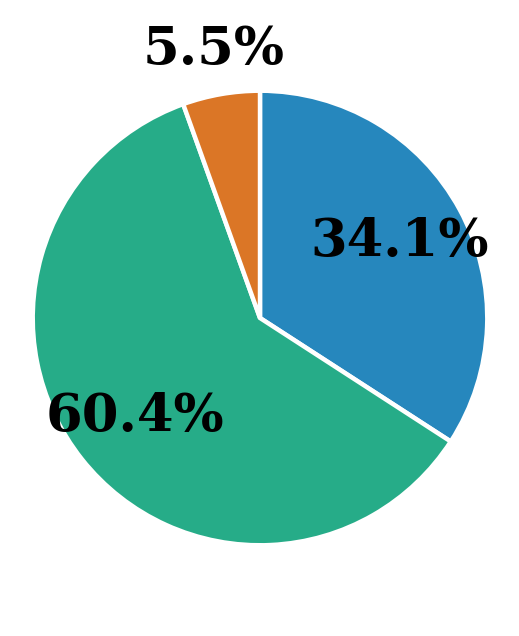}
        \caption{CoT}
        \label{fig:pie_cot}
    \end{subfigure}
    \hfill
    \begin{subfigure}[b]{0.28\linewidth}
        \includegraphics[width=\textwidth]{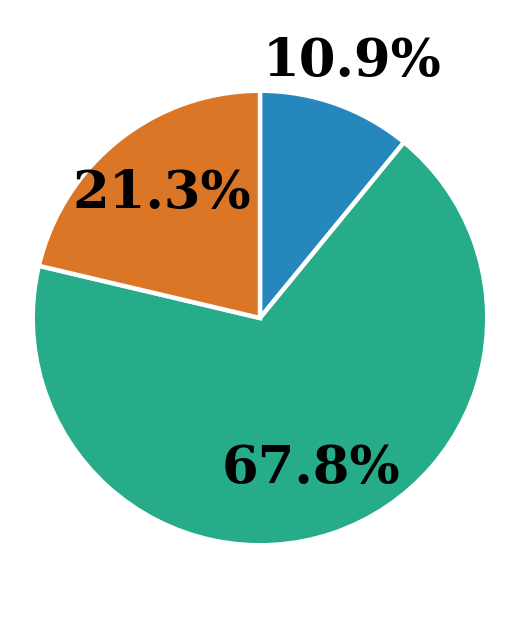}
        \caption{ToT}
        \label{fig:pie_tot}
    \end{subfigure}
    \hfill
    \begin{subfigure}[b]{0.28\linewidth}
        \includegraphics[width=\textwidth]{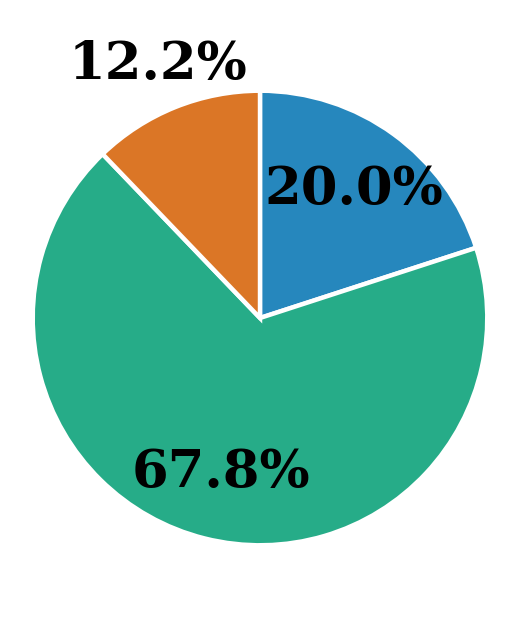}
        \caption{GoT}
        \label{fig:pie_got}
    \end{subfigure}

    \begin{subfigure}[b]{\linewidth}
        \includegraphics[width=.95\textwidth]{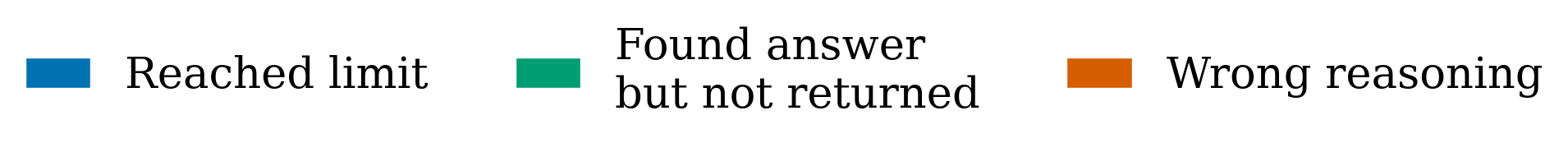}
        \label{fig:pie_legend}
    \end{subfigure}
    \caption{Error distribution across strategies. ToT and GoT reduce unanswered cases but increase logical errors due to more complex reasoning.}
    \label{fig:errors_pie}
    \vspace{-12pt}
\end{wrapfigure}

To understand failure patterns, we define three error types: (1) Reached limit — the reasoning hit the step limit; (2) Answer found but not returned — the correct answer appeared but was not output; (3) Wrong reasoning step — the model followed an illogical step. Using GPT-4o, we labeled a larger set of answers and traces. We observe in Figure~\ref{fig:errors_pie} that ToT and GoT show more “answer found but not returned” cases than CoT, suggesting better retrieval but occasional failures in synthesis. This comes with a slight rise in logical errors, likely due to the complexity of multiple reasoning paths.

\section{Limitations}
\label{sec:limitations}

In this work, we demonstrate how LLMs can be used to explore a graph while conditioning the next steps based on the graph's results. We show that the two approaches presented achieve superior results in graph exploration. Integrating KGs with LLMs can provide complex relational knowledge for LLMs to leverage. However, the overall effectiveness depends heavily on both the coverage and quality of the underlying graph, as well as the capabilities of the language model.

Extending inference-time reasoning methods for LLMs is significantly constrained by computational resources and the time available to the user. We analyze the computational complexity of the methods in Appendix \ref{app:Computational_Analysis}, where we show the exponential growth of ToT and GoT due to its branching structure. GoT further compounds this by allowing merges between reasoning paths, which increases the total number of evaluations. Additionally, loading large graphs into memory introduces substantial RAM overhead, limiting applicability to resource-rich environments.

While LLMs conditioned on external knowledge can generate outputs based on accessed content, their generated output is not strictly limited to that information. Thus, they may still hallucinate. Our framework helps mitigate this risk by grounding reasoning in explicit graph structure, but does not eliminate it.

\section{Conclusion}

We present a framework for grounding LLM reasoning in KGs by integrating each step with structured graph retrieval. By combining strategies like CoT, ToT, and GoT with adaptive graph search, our method achieves state-of-the-art performance on GRBench. Beyond performance, we find that explicitly linking reasoning steps to graph structure offers a more interpretable view of how LLMs navigate knowledge. The approach enables inference-time reasoning, offering flexibility across domains, and suggests a path toward reasoning interventions that are both systematic and transparent. Future work includes extending this framework to larger and more heterogeneous graphs, and exploring how structured retrieval can guide reasoning in domains where accuracy and verifiability are critical.

\paragraph{Disclaimers} 
This [paper/presentation] was prepared for informational purposes [“in part” if the work is collaborative with external partners]  by the Artificial Intelligence Research group of JPMorganChase and its affiliates ("JP Morgan'') and is not a product of the Research Department of JP Morgan. JP Morgan makes no representation and warranty whatsoever and disclaims all liability, for the completeness, accuracy or reliability of the information contained herein. This document is not intended as investment research or investment advice, or a recommendation, offer or solicitation for the purchase or sale of any security, financial instrument, financial product or service, or to be used in any way for evaluating the merits of participating in any transaction, and shall not constitute a solicitation under any jurisdiction or to any person, if such solicitation under such jurisdiction or to such person would be unlawful.

\copyright 2025 JPMorganChase. All rights reserved

% \section{Ethics Statement}

% This paper advances the integration of Knowledge Graphs and Large Language Models to enhance their capability for domain-specific question answering. This work can facilitate the development of chatbots that provide responses limited to certain answers, aiding in meeting industry-standard protocols and ensuring adherence to privacy guidelines.

% By anchoring the model’s reasoning processes in KGs, we aim to increase the transparency of LLM outputs. Each reasoning step is grounded in structured, retrievable information, enabling users to trace the origin of a given response. We admit that fully interpretable AI remains a challenge, and transparency remains a primary focus for better relying on such complex models.

\bibliographystyle{plainnat}
\bibliography{references}

%%%%%%%%%%%%%%%%%%%%%%%%%%%%%%%%%%%%%%%%%%%%%%%%%%%%%%%%%%%%

\appendix

\section{GRBench Statistics}
\label{app:grbench_statistics}

Detailed statistics of the graphs in GRBench \cite{graphCoT} are shown in Table \ref{Table:dataset_statistics}. \textbf{Academic Graphs} contain 3 types of nodes: paper, author, venue. \textbf{Literature Graphs} contain 4 types of nodes: book, author, publisher and series. \textbf{Healthcare Graph} contains 11 types of nodes: anatomy, biological process, cellular component, compound, disease, gene, molecular function, pathway, pharmacologic class, side effect, and symptom. Questions are created according to multiple templates labeled as easy, medium, and hard, depending on the number of nodes required to give the answer.

\begin{table}[h]
\centering
\resizebox{0.73\textwidth}{!}{
\begin{tabular}{cccccc}
\toprule
\multirow{2}{*}{\textbf{Domain}}   & \multirow{2}{*}{\textbf{Topic}} & \multicolumn{2}{c}{\textbf{Graph Statistics}} & \multicolumn{2}{c}{\textbf{Data}}    \\ \cline{3-6} 
                          &                        & \textbf{\# Nodes}          & \textbf{\# Edges }        & \textbf{\# Templates} & \textbf{\# Questions} \\
\hline
\multirow{5}{*}{Academic} & Biology                & $\sim$4M          & $\sim$39M        & 14           & 140          \\ \cline{2-6} 
                          & Chemistry              & $\sim$4M          & $\sim$30M        & 14           & 140          \\ \cline{2-6} 
                          & Material Science       & $\sim$3M          & $\sim$22M        & 14           & 140          \\ \cline{2-6} 
                          & Medicine               & $\sim$6M          & $\sim$30M        & 14           & 140          \\ \cline{2-6} 
                          & Physics                & $\sim$2M          & $\sim$33M        & 14           & 140          \\
\hline
Literature                & Goodreads              & $\sim$3M          & $\sim$22M        & 24           & 240          \\
\hline
Healthcare                & Disease                & $\sim$47K         & $\sim$4M         & 27           & 270          \\
\hline
\cellcolor{gray!25} \textbf{SUM}                       & \cellcolor{gray!25} -                      & \cellcolor{gray!25} -                 & \cellcolor{gray!25} -                &  \cellcolor{gray!25} \textbf{121}          & \cellcolor{gray!25} \textbf{1210}       \\
\bottomrule
\end{tabular}
}
\caption{Detailed statistics of the GRBench \cite{graphCoT}.}
\label{Table:dataset_statistics}
\end{table}

\section{LLM <--> KG Interaction Pipelines}
\label{App:pipelines}

\begin{figure}[h]
    \centering
    \includegraphics[width=0.8\linewidth]{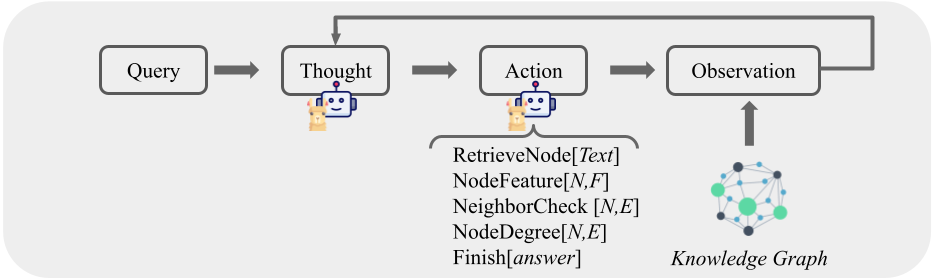}
    \caption{Agent Pipeline: (1) Input Query, (2) Thought Generation (3) Action Selection, (4) Environment Observation from the Knowledge Graph. The process is repeated until termination action is generated or limit reached.}
    \label{fig:agent_pipeline}
\end{figure}

\begin{figure}[h]
    \centering
    \includegraphics[width=0.8\linewidth]{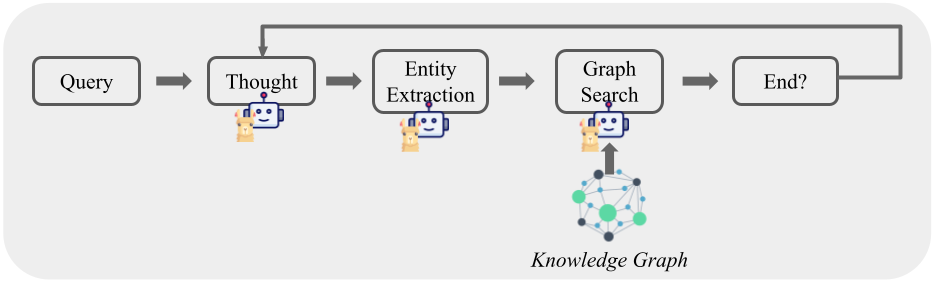}
    \caption{Automatic Graph Exploration Pipeline: (1) Input Query, (2) Thought Generation, (3) Entity Extraction (from query or thought with LLM), (4) Automatic Graph Search as described in Algorithm \ref{alg:graph_exploration_algorithm} (5) Query LLM for answer or continue}
    \label{fig:graph_explore_pipeline}
\end{figure}

Description of the two LLM + KG Interaction Pipelines in their CoT form:
\begin{enumerate}
    \item Agent -- Figure \ref{fig:agent_pipeline}A pipeline where the LLM alternates between generating a reasoning step, selecting an explicit action (e.g., retrieving a node, checking neighbors), and observing results from the KG until termination.
    \item Automatic Graph Exploration -- Figure \ref{fig:graph_explore_pipeline}. A pipeline where entities are automatically extracted from the LLM’s generated text and used to guide iterative graph traversal with pruning, progressively expanding the reasoning chain.
\end{enumerate}

\newpage

Algorithm \ref{alg:graph_exploration_algorithm} presents the pseudocode for the Automatic Graph Exploration described in Section \ref{Section:Automatic_Graph_Exploration}.

\begin{algorithm}
\setstretch{0.95}
\caption{Graph Exploration Algorithm}
\label{alg:graph_exploration_algorithm}
\begin{algorithmic}[1]
\Procedure{GraphExplore}{$LLM, seen\_entities, search\_depth$}
    \State $relevant\_attributes$, $found\_triples \gets 0$  
    \For{$depth$ in $search\_depth$}
        \For{$entity$ in $seen\_entities$}
            \If{$seen\_entities[entity\_id].visited$ == True}
                \State Continue
            \Else
                \State $seen\_entities[entity] \gets$ Visited
            \EndIf
            \State $head\_entity\_name$, $entity\_attributes$, $neighbors \gets$  Graph[$entity$]
            \State $pruned\_neighbors \gets$ prune\_relations($LLM, neighbors$)
            \State $pruned\_neighbors \gets$ prune\_entities($LLM, pruned\_neighbors$)
            \State $found\_triples \gets$ generate\_triples($entity, pruned\_neighbros$)
        \EndFor
        \State $seen\_entities \gets$ Update($seen\_entities$, neighbors)
        \If{End?($LLM, found\_triples, relevant\_attributes$) == True}
            \State break
        \EndIf
    \EndFor
    \State \Return $found\_triples, relevant\_attributes, seen\_entities$
\EndProcedure

\end{algorithmic}

\end{algorithm}

\section{Performance results in plots}

\begin{figure}[h]
    \centering
    \includegraphics[width=.96\linewidth]{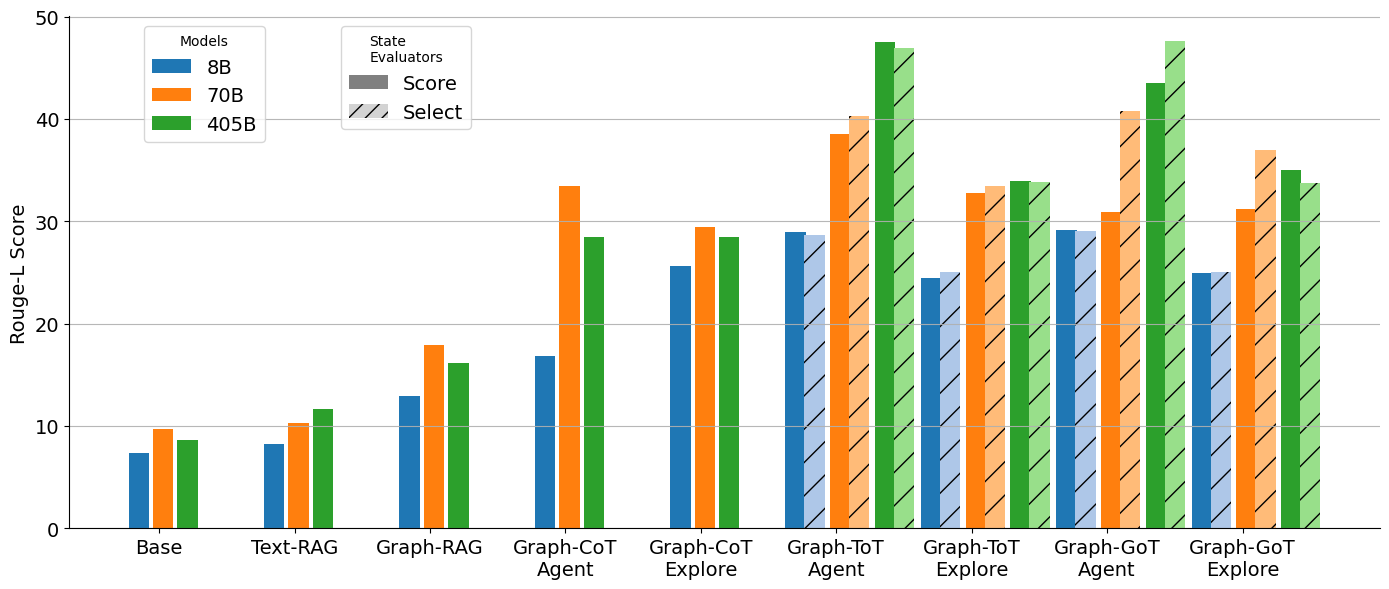}
    \caption{Performance results using the Rouge-L metric on the healthcare graph of GRBench \cite{graphCoT}, comparing all methods with LLama 3.1 Instruct models of various sizes (8B, 70B, 405B-FP8). Experimental details are included in Section \ref{Section:Experiments}.}
    \label{fig:plot_result_rougeL_healthcare}
\end{figure}

\begin{figure}[h]
    \centering
    \includegraphics[width=\linewidth]{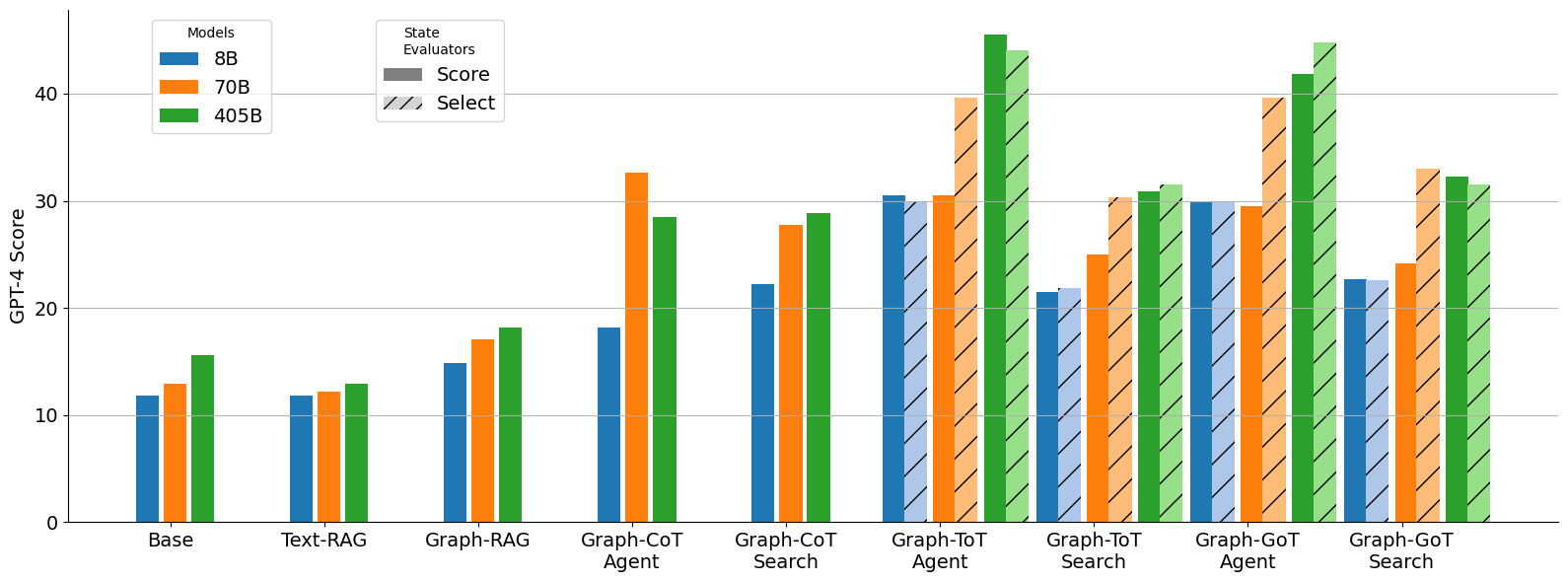}
    \caption{Performance results using the GPT4Score metric on the healthcare graph of GRBench \cite{graphCoT}, comparing all methods with LLama 3.1 Instruct models of various sizes (8B, 70B, 405B-FP8). Experimental details are included in Section \ref{Section:Experiments}.}
    \label{fig:plot_result_gpt4score_healthcare}
\end{figure}

Figures \ref{fig:plot_result_rougeL_healthcare} and \ref{fig:plot_result_gpt4score_healthcare} illustrate the performance results using the Rouge-L and GPT4score metrics, respectively, for the healthcare graph for all methods. The results were run on the LLama 3.1 Instruct models (8B, 70B, and 405B-FP8) and demonstrate the improved performance achieved through more complex reasoning and search strategies during inference.

\newpage

\section{Results on GPT4Score}
\label{App:results-gpt4score}

In this section, we present the results of the experiments described in Section \ref{Section:Experiments} for all methods, using the GPT4Score metric. This metric calculates the percentage of "correct" answers as judged by GPT-4 when presented with both the correct and the generated answer. The tables in this section present the same data as in Tables \ref{Table:Results} and \ref{Table:Results_GoT}, but evaluated using GPT4Score.

\begin{table}[h]
\centering
\resizebox{.9\textwidth}{!}{
\begin{tabular}{llllccccccc}
\toprule
& \textbf{Method} & & \multicolumn{1}{c}{\textbf{Model}} & \textbf{Healthcare} & \textbf{Goodreads} & \textbf{Biology} & \textbf{Chemistry} & \textbf{Materials Science} & \textbf{Medicine} & \textbf{Physics} \\
                  \hline
\multirow{9}{*}{\rotatebox{90}{\shortstack{Baselines}}} & \multicolumn{2}{l}{}                  & Llama 3.1 8B-Ins                   & 11.85                                & 13.33             & 10.71            & 11.43              & 7.86                      & 7.87            & 9.29            \\
& \multicolumn{2}{l}{Base}                  & Llama 3.1 70B-Ins                   & 12.96                                 & 19.17               & 10.00            & 12.14              & 11.43                       & 11.43             & 12.86            \\
& \multicolumn{2}{l}{}                  & Llama 3.1 405B-Ins                   & 15.55                               & 26.67              & 12.86           & 12.14              & 12.14                     & 13.57            & 12.14            \\

\cdashline{2-11}
& \multicolumn{2}{l}{}                  & Llama 3.1 8B-Ins                   & 11.85                               & 21.67              & 12.86            & 10.00              & 10.00                      & 8.57             & 7.86            \\
                  & \multicolumn{2}{l}{Text-RAG}                  & Llama 3.1 70B-Ins                   & 12.22                                & 27.5              & 12.14            & 13.57             & 13.57                      & 13.57             & 12.86            \\
                  & \multicolumn{2}{l}{}                  & Llama 3.1 405B-Ins           & 12.96                                & 26.67              & 15.00           & 13.57              & 12.86                      & 14.29           & 13.57            \\ \cdashline{2-11}
                  & \multicolumn{2}{l}{}                  & Llama 3.1 8B-Ins                   & 14.81                               & 32.50             & 29.29               & 29.28              & 27.86                       & 25.71             & 29.29            \\
                  & \multicolumn{2}{l}{Graph-RAG}                  & Llama 3.1 70B-Ins                   & 17.04                                & 32.92              & 39.29               & 40.71              & 43.57                      & 34.29             & 40.00    \\
                  & \multicolumn{2}{l}{}                  & Llama 3.1 405B-Ins         & 18.15                                & 31.67             & 37.14               & 42.86              & 40.00                       & 36.43             & 41.43 \\
                  \hline \hline
 \multirow{6}{*}{\rotatebox{90}{\shortstack{Graph\\CoT}}} & \multicolumn{2}{l}{}                  & Llama 3.1 8B-Ins                   & 18.15                                & 32.5               & 20.71            & 19.28               & 25.00                      & 14.29            & 21.43            \\
 & \multicolumn{2}{l}{Agent}                  & Llama 3.1 70B-Ins                   & 32.59                               & 43.75              & 50.00            & 51.43              & 50.00                      & 48.57            & 46.43           \\ 
 & \multicolumn{2}{l}{}                  & Llama 3.1 405B-Ins                  & 28.89                              & 48.33              & 38.57            & 38.57             & 47.86                     & 56.43            & 34.29           \\ \cdashline{2-11}
                  & \multicolumn{2}{l}{ \multirow{3}{*}{\shortstack{Graph\\Explore}}}                    & Llama 3.1 8B-Ins                   & 22.22                                & 36.67              & 35.00            & 30.71              & 29.29                      & 29.29                & 32.86            \\
                &  \multicolumn{2}{l}{}              & Llama 3.1 70B-Ins                   & 27.78                                & 32.92            & 45.71          & 49.29             & 40.00                     & 40.00             & 44.29            \\
                & \multicolumn{2}{l}{}                  & Llama 3.1 405B-Ins                   & 28.89                               & 48.33            & 38.57       & 38.57              & 47.86                      & 56.43                & 34.29            \\
                  \hline \hline
                  
\multirow{12}{*}{\rotatebox{90}{\shortstack{Graph\\ToT}}} & \multirow{6}{*}{Agent} &    \multirow{3}{*}{Score}          & Llama 3.1 8B-Ins                  & 30.49                                & 55.14              & 43.33            & 41.67               & 44.05                      & 36.43             & 39.52            \\
& &              & Llama 3.1 70B-Ins                  & 30.49                                & 54.48              & 65.48            & 62.14              & 55.95                      & 63.57             & 56.19            \\
& &              & Llama 3.1 405B-Ins                  & \textbf{45.55}                                & 56.53              & \textbf{71.67}            & 65.71               & 52.62                      & 68.81             & 44.76            \\ \cdashline{3-11}

    &                   &        \multirow{3}{*}{Select}           & Llama 3.1 8B-Ins                   & 30.00                                 & 54.17               & 40.71            & 37.14              & 40.00                      & 32.86             & 36.43            \\
    &                   &                  & Llama 3.1 70B-Ins                   & 39.63                                 & 56.67               & 65.00            & 67.14              & 62.86                      & 60.71             & 55.55            \\

                  &                   &                   & Llama 3.1 405B-Ins                   & 44.07                                 & \textbf{58.75}              & 71.43            & \textbf{69.29}              & \textbf{65.00}                      & \textbf{68.81}             & \textbf{60.00}            \\  \cdashline{2-11}
                  & \multirow{6}{*}{\shortstack{Graph\\Explore}} &      \multirow{3}{*}{Score}             & Llama 3.1 8B-Ins                   & 21.48                                & 41.10              & 32.86            & 31.67              & 31.43                      & 32.14             & 35.24            \\ 
                  &                   &                   & Llama 3.1 70B-Ins                   & 24.94                                & 40.97              & 52.38            & 57.86              & 49.29                      & 54.29                & 47.86           \\
                  &                   &                   & Llama 3.1 405B-Ins                   & 30.86                                & 48.33              & 47.86            & 57.14              & 50.71                      & 56.67                & 47.14           \\ \cdashline{3-11}

                  &                   &     \multirow{3}{*}{Select}              & Llama 3.1 8B-Ins   &      21.85    &     41.67                      &   32.86            & 31.67            & 31.43              & 32.14                & 35.24           \\
                  &                   &                   & Llama 3.1 70B-Ins                   & 30.37                                & 42.08              & 54.29            & 57.14              & 47.86                      & 52.14                & 46.43           \\
                  &                   &                   & Llama 3.1 405B-Ins                   & 31.48                                & 48.75              & 45.00            & 57.86              & 48.86                      & 57.14                & 45.71           \\
\bottomrule
\end{tabular}
}
\caption{GPT4Score performance results on GRBench \cite{graphCoT}, comparing standard LLMs, Text-RAG, Graph-RAG, Graph-CoT, and Graph-ToT. Experiments are described in Section \ref{Section:Experiments}, using LLama 3.1 - Instruct backbone models with sizes 8B, 70B, and 405B.}

\label{Table:Results-GPT4Score}
\end{table}

\begin{figure}[h]
    \centering
    \begin{minipage}{0.48\textwidth}
        \centering
        \resizebox{0.8\textwidth}{!}{
            \begin{tabular}{cclcc}
                \toprule
                \multicolumn{2}{c}{\textbf{Method}}                        & \multicolumn{1}{c}{\textbf{Model}}              & \textbf{Healthcare} & \textbf{Biology} \\
                \hline
                \multirow{6}{*}{\rotatebox{90}{Agent}}  & \multirow{3}{*}{\rotatebox{90}{Score}}  & Llama 3.1 8B-Ins   & 29.88           &    32.86                                                     \\
                                                &                         & Llama 3.1 70B-Ins  &   29.51         &    61.69                                                     \\
                                                &                         & Llama 3.1 405B-Ins &    41.81        &   48.33                                                     \\
                                                \cdashline{2-5}
                                                & \multirow{3}{*}{\rotatebox{90}{Select}} & Llama 3.1 8B-Ins   &  30.00          & 40.71                                                         \\
                                                &                         & Llama 3.1 70B-Ins  &  39.63          &     69.83                                                    \\
                                                &                         & Llama 3.1 405B-Ins &   \textbf{44.81}        &   \textbf{72.86}                                                      \\
                                                \hline
                \multirow{6}{*}{\rotatebox{90}{Graph Explore}} & \multirow{3}{*}{\rotatebox{90}{Score}}  & Llama 3.1 8B-Ins   &   22.72         &      21.19                                                   \\
                                                &                         & Llama 3.1 70B-Ins  &   24.20         &     48.57                                                    \\
                                                &                         & Llama 3.1 405B-Ins & 32.22        &   41.67                                                      \\
                                                \cdashline{2-5}
                                                & \multirow{3}{*}{\rotatebox{90}{Select}} & Llama 3.1 8B-Ins   &  22.59          &  19.28                                                       \\
                                                &                         & Llama 3.1 70B-Ins  &   32.96         &      52.86                                                   \\
                                                &                         & Llama 3.1 405B-Ins &   31.48         &   57.86
                                                \\
                \bottomrule
            \end{tabular}
        }
        \caption{Graph-GoT results (GPT4Score) on GRBench  with Llama 3.1 Instruct sizes 8B, 70B, and 405B.}
        \label{Table:Results_GoT_GP4Score}
    \end{minipage}
    \hfill
    \begin{minipage}{0.48\textwidth}
        \centering
        \includegraphics[width=.65\textwidth]{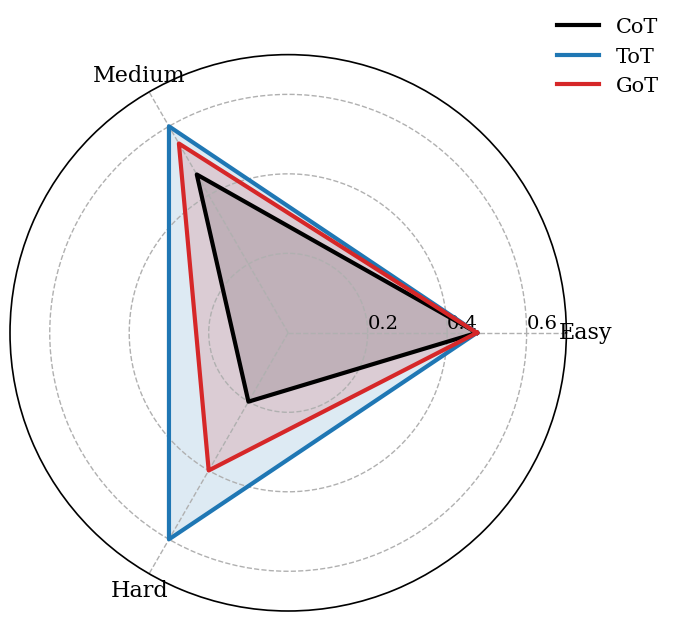}
        \caption{Results decomposed into easy/medium/hard questions according to GPT4Score.}
        \label{fig:radar_levels}
    \end{minipage}
\end{figure}

\newpage

\section{Computational Analysis}
\label{app:Computational_Analysis}

\begin{wrapfigure}{r}{0.38\textwidth}
    \centering
    \includegraphics[width=\linewidth]{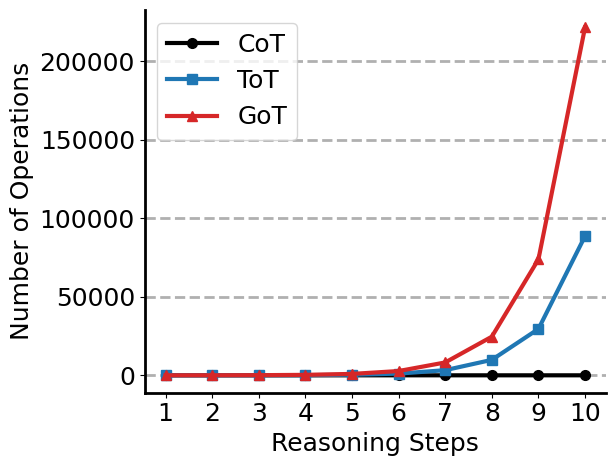}
    \caption{Comparison of computational costs across different reasoning strategies — CoT, ToT, GoT — as a function of reasoning steps. GoT exhibits the highest cost due to merge operations and graph traversal.}
    \label{fig:ops-steps}
\end{wrapfigure}

\begin{table}[h]
\centering
\small
\resizebox{0.95\textwidth}{!}{
\begin{tabular}{@{}c p{2.0cm} p{1.7cm} p{3cm} p{4.0cm} p{2.5cm}@{}}
\toprule
& \textbf{Method} & \textbf{Key Parameters} & \textbf{Approx. \# LLM Calls} & \textbf{Approx. \# KG Operations} & \textbf{Primary Growth Driver(s)} \\
\midrule

\multirow{2}{*}{\rotatebox[origin=c]{90}{CoT}} 
& Agent & $n$ & $\mathcal{O}(n)$ & $\mathcal{O}(n)$ & $n$ (linear) \\
\cmidrule(lr){2-6}
& Graph Explore & $n$, $d$ & $\mathcal{O}(n)$ & $\mathcal{O}(n \cdot \text{Cost}_{\text{Explore}}(d))$ & $n$, $d$ \\

\midrule

\multirow{2}{*}{\rotatebox[origin=c]{90}{ToT}} 
& Agent & $D_{\text{max}}$, $k$, $t$ & $\mathcal{O}\left(k \cdot \frac{t^{D_{\text{max}}} - 1}{t - 1}\right)$ & Same as LLM Calls & $D_{\text{max}}$, $k$, $t$ (exponential in $D_{\text{max}}$) \\
\cmidrule(lr){2-6}
& Graph Explore & $D_{\text{max}}$, $k$, $t$, $d$ & $\mathcal{O}\left(k \cdot \frac{t^{D_{\text{max}}} - 1}{t - 1}\right)$ & $\mathcal{O}\left(k \cdot \frac{t^{D_{\text{max}}} - 1}{t - 1} \cdot \text{Cost}_{\text{Explore}}(d)\right)$ & $D_{\text{max}}$, $k$, $t$, $d$ \\

\midrule

\multirow{2}{*}{\rotatebox[origin=c]{90}{GoT}} 
& Agent & $D_{\text{max}}$, $k$, $t$ & {\tiny     .     $\mathcal{O}\left(k \cdot \frac{t^{D_{\text{max}}} - 1}{t - 1} + \sum_{i=1}^{D_{\text{max}}} \left\lfloor \frac{k \cdot t^i}{2} \right\rfloor \right)$} & Same as LLM Calls & $D_{\text{max}}$, $k$, $t$ (aggregation adds extra cost) \\
\cmidrule(lr){2-6}
& Graph Explore & $D_{\text{max}}$, $k$, $t$, $d$ & Same as Agent & $\mathcal{O}\left(\text{LLM Calls} \cdot \text{Cost}_{\text{Explore}}(d)\right)$ & $D_{\text{max}}$, $k$, $t$, $d$ \\
\bottomrule
\end{tabular}
}
\caption{Theoretical computational complexity comparison of reasoning methods. Parameters: $n$ (reasoning steps), $D_{\text{max}}$ (tree depth), $k$ (branching factor), $t$ (paths retained), $d$ (KG search depth), and $\text{Cost}_{\text{Explore}}(d)$ (cost per KG search). GoT includes pairwise aggregation of thoughts at each depth.}
\label{tab:complexity_comparison}
\end{table}

The computational analysis summarized in Table~\ref{tab:complexity_comparison} highlights the trade-offs between reasoning power and computational cost when grounding LLMs with Knowledge Graphs (KGs). The \textit{Agent}-based methods scale linearly with the number of reasoning steps or tree nodes, with CoT representing the lowest cost baseline. In contrast, \textit{Automatic Graph Exploration} methods introduce additional overhead via entity extraction, multi-hop traversal up to a fixed \texttt{max\_depth}, and LLM-based pruning at each step.

Among reasoning strategies, ToT introduces exponential growth in cost with respect to depth due to its exploration of $k$ branches and selection of $t$ continuations per level. GoT further amplifies this by incorporating aggregation transformations that attempt to merge every pair of thoughts at each depth, leading to an additional cost proportional to $\sum_{i=1}^{D_{\text{max}}} \left\lfloor \frac{k \cdot t^i}{2} \right\rfloor$.

Importantly, our experiments reveal that the higher complexity of GoT does not consistently translate to improved accuracy compared to ToT, suggesting diminishing returns. While the model size (e.g., 8B, 70B, 405B) influences the latency and memory footprint of each LLM call, it does not affect the algorithmic complexity classes shown. Hence, selecting a strategy requires balancing reasoning depth with feasible compute budgets.

\section{Prompts Archive}

In this section, we gather the necessary prompts essential for implementing the proposed methodologies.

\begin{tcolorbox}[breakable, colframe=black, title=Agent]
\texttt{
\textbf{Agent Step}: Solve a question answering task with interleaving Thought, Interaction with Graph, Feedback from Graph steps. In Thought step, you can think about what further information is needed, and In Interaction step, you can get feedback from graphs with four functions:\\
(1) RetrieveNode[keyword], which retrieves the related node from the graph according to the corresponding query.\\
(2) NodeFeature[Node, feature], which returns the detailed attribute information of Node regarding the given "feature" key.
(3) NodeDegree[Node, neighbor\_type], which calculates the number of "neighbor\_type" neighbors of the node Node in the graph.\\
(4) NeighbourCheck[Node, neighbor\_type], which lists the "neighbor\_type" neighbours of the node Node in the graph and returns them.\\
You may take as many steps as necessary.\\
Here are some examples:\\
\{examples\}\\
Please answer by providing node main feature (e.g., names) rather than node IDs.\\
Generate the next step.\\
Definition of the graph: \{graph\_definition\}\\
Question: \{question\}\\
\{scratchpad\}}
\end{tcolorbox}

\begin{tcolorbox}[breakable, colframe=black, title=Automatic Graph Exploration]
\texttt{\textbf{Search Thought}: Given the previous thoughts, generate the next thought to answer the provided question.\\
Your end goal is to answer the question step by step. 
For context, you are also provided with some knowledge triples from a knowledge base. \\
Follow the format of the examples to generate the next thought.\\
\\
\{examples\}\\
\\
Graph Definition: \{graph\_definition\}\\
Question: \{question\}\\
Knowledge Triples: \\
    \{triples\}\\
Previous thoughts:\\
    \{thoughts\}\\
Related Entity Attributes:\\
    \{attributes\}\\
Next Thought:\\
\\
\textbf{Search End?}: Your are provided with the an original question, the associated subquestion thoughts and their corresponding knowledge graph triples (head\_entity -> relation -> tail\_entity). 
Your task is to answer whether it's sufficient for you to answer the original question (Yes or No).
You are provided with examples. You should follow the same format as in the examples, writing 'Yes' or 'No' within brackets at the beginning of the answer.\\
\textit{(Examples)\\
Task: 
Question: \{\textit{question}\}\\
Thoughts: \{\textit{thoughts}\}\\
Knowledge Triples: \{\textit{triples}\}\\
Entity Attributes: \{\textit{attributes}\}\\
Answer: \\
\\
\textbf{Entity Extraction}: Given the provided text, extract the relevant entities that may appear in a knowledge base.
Return the answer at the end with brackets {{relevant entities}} as shown in the following examples. If there are several entities, separate them with commas.\\
(\textit{Examples})\\
Task:
Text: \{\textit{text}\}\\
Relevant Entities: \\
\\
\textbf{Prune Relations}: From the given entity and relations, select only the relevant relations to answer the question.
Provide the answer at the end with brackets{{answer}} , as shown in the following example.\\
(\textit{Examples})\\
Question: \{\textit{question}\}\\
Head Entity: \{\textit{entity}\}\\
Relations: \{\textit{relations}\}
Answer:\\
\\
\textbf{Prune entities}: You are provided with a question, a head entity, a relation and tail entity or entities from a knowledge base.
Select the tail entity or entities to answer the question.
Return the tail entity or entities at the end with brackets {{relevant entity or entities}}, as shown in the following examples.\\
(\textit{Examples})\\
Question: \{\textit{question}\}\\
Head Entity: \{\textit{head\_entity}\}\\
Relation: \{\textit{relation}\}\\
Tail Entities: \{\textit{tail\_entities}\}\\
Relevant Entities: \\
\\
\textbf{Search Attributes}: Is any of the attributes relevant to answer the question?
Return the answer at the end with brackets {{answer}} , as shown in the following examples.\\
(\textit{Examples})\\
Question: \{\textit{question}\}\\
Entity: \{\textit{entity}\}\\
Attributes: \{\textit{attributes}\}\\
Relevant Attributes: 
}
}
\end{tcolorbox}

\begin{tcolorbox}[breakable, colframe=black, title=State Evaluators]
\textbf{Selection Vote}: Given a question, you need to select the possible chain of thought that may lead to the correct answer with higher probablity. 
You are provided with several choices with thouhgts and related triples from a knowledge base. Decide which choice is most promising to complete the task.
Analyze each choice in detail, then conclude in the last line:
"The best choice is {{s}}" , where s the integer id of the choice. \\
(\textit{Examples})\\
Question: \{\textit{question}\}\\
Choices: \{\textit{choices}\} (+ attributes/triples)\\
Answer: \\
\\
\textbf{Score Vote}: Generate a score for the given reasoning chain. 
The score represents the probability that the chain will lead to the correct answer.
The chains contain interleaved thoughts and related triples from a knowledge base.
Some chains may not be complete, but you need to judge the steps that are provided.
The score can be any floating number between 0 and 1.\\
(\textit{Examples})\\
Question: \{\textit{question}\}\\
Thought Chain: \{\textit{thoughts}\} (+ attributes/triples)\\
Score: 
\end{tcolorbox}

\begin{tcolorbox}[breakable, colframe=black, title=Graph of Thoughts]
\textbf{Graph of Thought Merge}: Generate the next thought for the merged chain of thoughts.
You are provided with the question, two chains of thoughts, and the corresponding merged chain of thought. 
Identify inconsistencies or errors from the previous chains and provide the next thought for the merged chain.
You should follow the same format as in the examples.\\
(\textit{Examples})\\
Question: \{\textit{question}\}\\
Chain 1: \{\textit{chain\_1}\} (+triples/attributes)\\
Chain 2: \{\textit{chain\_2}\} (+triples/attributes)\\
Merged Chain: \{\textit{merged\_chain}\}\\
Next Thought:
\end{tcolorbox}

\newpage

\section{Examples}

\begin{table*}[h]

\renewcommand{\arraystretch}{1.2}
\setlength{\tabcolsep}{8pt}
\begin{tabularx}{\textwidth}{@{}p{.48\textwidth} p{.48\textwidth}@{}}
\textbf{Example 1: Multi-step Reasoning over Graph} & \textbf{Example 2: Direct Triple Lookup} \\

\textbf{Question:} \textit{What anatomy can be expressed by gene KRT39?} \newline
\textbf{Ground Truth Answer:} \textit{head, skin of body} \newline
\textbf{Model Answer:} \textit{head, skin of body} \newline

\begin{minipage}[t]{\linewidth}
\textbf{Thought 1:} The question is related to a gene node (KRT39). We need to find this node in the graph. \\
\textbf{Action 1:} \texttt{RetrieveNode[KRT39]} \\
\textbf{Observation 1:} The ID of the node is \texttt{390792}.

\vspace{0.5em}
\textbf{Thought 2:} We need to check the 'Anatomy-expresses-Gene' neighbors of this gene node. \\
\textbf{Action 2:} \texttt{NeighbourCheck[390792, Anatomy-expresses-Gene]} \\
\textbf{Observation 2:} The neighbors are \texttt{['UBERON:0000033', 'UBERON:0002097']}.

\vspace{0.5em}
\textbf{Thought 3:} Retrieve names of the anatomy nodes. \\
\textbf{Action 3:} \texttt{NodeFeature[UBERON:0000033, name]}, \texttt{NodeFeature[UBERON:0002097, name]} \\
\textbf{Observation 3:} \texttt{UBERON:0000033} $\rightarrow$ \textit{head}, \texttt{UBERON:0002097} $\rightarrow$ \textit{skin of body}.

\vspace{0.5em}
\textbf{Thought 4:} These are the anatomy terms expressed by the gene. \\
\textbf{Action 4:} \texttt{Finish[head, skin of body]}
\end{minipage}
&
\textbf{Question:} \textit{What anatomy can be expressed by gene KRT39?} \newline
\textbf{Ground Truth Answer:} \textit{head, skin of body} \newline
\textbf{Model Answer:} \textit{head, skin of body} \newline

\begin{minipage}[t]{\linewidth}
\textbf{Thought 1:} \texttt{KRT39} is a gene that is known to be expressed in two anatomical regions.

\textbf{Found Triples:}
\begin{itemize}[leftmargin=*]
    \item \texttt{"KRT39" --> Anatomy-expresses-Gene --> head}
    \item \texttt{"KRT39" --> Anatomy-expresses-Gene --> skin of body}
\end{itemize}

\textbf{Action:} \texttt{Finish[head, skin of body]}
\end{minipage}
\\
\end{tabularx}
\caption{Comparison of two LLM-generated reasoning traces for the same question using different strategies: multi-step graph traversal (Example 1) vs. direct triple lookup (Example 2).}
\end{table*}

\end{document}